\documentclass{article}

\usepackage{microtype}
\usepackage{graphicx}
\usepackage{subfigure}
\usepackage{booktabs} % for professional tables

\usepackage{hyperref}

\usepackage[accepted]{icml2024}

\usepackage{amsmath}
\usepackage{amssymb}
\usepackage{mathtools}
\usepackage{amsthm}

\usepackage[capitalize,noabbrev]{cleveref}

\theoremstyle{plain}
\newtheorem{theorem}{Theorem}[section]
\newtheorem{proposition}[theorem]{Proposition}

\theoremstyle{definition}

\theoremstyle{remark}

\usepackage{amsmath,amsfonts,bm}

\def\eqref#1{equation~\ref{#1}}
\def\1{\bm{1}}

\DeclareMathAlphabet{\mathsfit}{\encodingdefault}{\sfdefault}{m}{sl}
\SetMathAlphabet{\mathsfit}{bold}{\encodingdefault}{\sfdefault}{bx}{n}

\usepackage{algorithm}

\usepackage{multirow}
\usepackage{makecell}
\usepackage{wrapfig}
\newcommand{\minsu}[1]{{\color{black}#1}}

\newcommand{\ours}{SRT}

\icmltitlerunning{Symmetric Replay Training}

\begin{document}

\twocolumn[
\icmltitle{Symmetric Replay Training: Enhancing Sample Efficiency\\in Deep Reinforcement Learning for Combinatorial Optimization}




\begin{icmlauthorlist}
\icmlauthor{Hyeonah Kim}{kaist}
\icmlauthor{Minsu Kim}{kaist}
\icmlauthor{Sungsoo Ahn}{postech}
\icmlauthor{Jinkyoo Park}{kaist,omelet}
\end{icmlauthorlist}

\icmlaffiliation{kaist}{Department of Industrial \& Systems Engineering, Korea Advanced Institute of Science and Technology (KAIST), Daejeon, South Korea}
\icmlaffiliation{postech}{Pohang University of Science and Technology (POSTECH), Pohang, South Korea}
\icmlaffiliation{omelet}{OMELET, Daejeon, South Korea}

\icmlcorrespondingauthor{Hyeonah Kim}{hyeonah\_kim@kaist.ac.kr}

\icmlkeywords{Combinatorial Optimization, Sample Efficiency, Replay Training, Deep Reinforcement Learning, Imitation Learning}

\vskip 0.3in
]



\printAffiliationsAndNotice{}  % leave blank if no need to mention equal contribution

\begin{abstract}
Deep reinforcement learning (DRL) has significantly advanced the field of combinatorial optimization (CO). However, its practicality is hindered by the necessity for a large number of reward evaluations, especially in scenarios involving computationally intensive function assessments. To enhance the sample efficiency, we propose a simple but effective method, called \textit{symmetric replay training (SRT)}, which can be easily integrated into various DRL methods.
Our method leverages high-reward samples to encourage exploration of the under-explored symmetric regions without additional online interactions -- \emph{free}. Through replay training, the policy is trained to maximize the likelihood of the symmetric trajectories of discovered high-rewarded samples. 
Experimental results demonstrate the consistent improvement of our method in sample efficiency across diverse DRL methods applied to real-world tasks, such as molecular optimization and hardware design. 
\end{abstract}

\section{Introduction} \label{sec:intro}
Combinatorial optimization (CO) problems arise across diverse industrial domains, but they are notoriously challenging to solve. In CO (e.g., traveling salesman problems; TSP), the massive discrete solution space often leads to NP-hardness. Furthermore, CO problems in practical scenarios often involve computationally expensive objective functions to evaluate (e.g., a black-box function), introducing significant restrictions on the problem-solving process. Even though solving CO problems with expensive objective functions is frequently found in various fields like drug discovery or hardware design, it has numerous challenges.

Recent advances in deep reinforcement learning (DRL) have drawn significant attention as an alternative way to solving CO problems. Various studies have demonstrated remarkable achievements across broad domains, including routing \citep{kool2018attention, kwon2020pomo, son2023meta}, scheduling \citep{park2021schedulenet, kwon2021matrix}, and material design \citep{olivecrona2017molecular,ahn2020guiding,bengio2021gflownet}. 
Especially in constructive methods, where a solution is sequentially built from an empty set, it is beneficial to guarantee feasibility to practical constraints using a masking strategy. Note that in the constructive method, a policy generates a trajectory, and the reward is evaluated in a terminal state, i.e., episodic rewards.

DRL methods generally leverage extensive samples to enhance training stability, assuming reward evaluation is affordable. However, in practice, reward computation can be costly due to several factors: (1) objective functions often cannot be defined analytically, requiring computationally intensive simulations for evaluation, such as in black-box optimization; or (2) objective functions may be analytically defined but still demand additional computation, such as solving another optimization problem (e.g., bi-level optimization). For instance, in molecular optimization, evaluating designed molecules in real-world scenarios involves costly assessments, such as wet lab experiments \citep{gao2022sample}.
In this regard, reducing the number of reward evaluations is significantly beneficial. This discrepancy raises a straightforward research question: \emph{How can we enhance the sample efficiency of DRL methods for a broad spectrum of CO problems and methods?}

\begin{figure*}
    \centering
    % \vspace{-3pt}
    \includegraphics[width=0.85\textwidth]{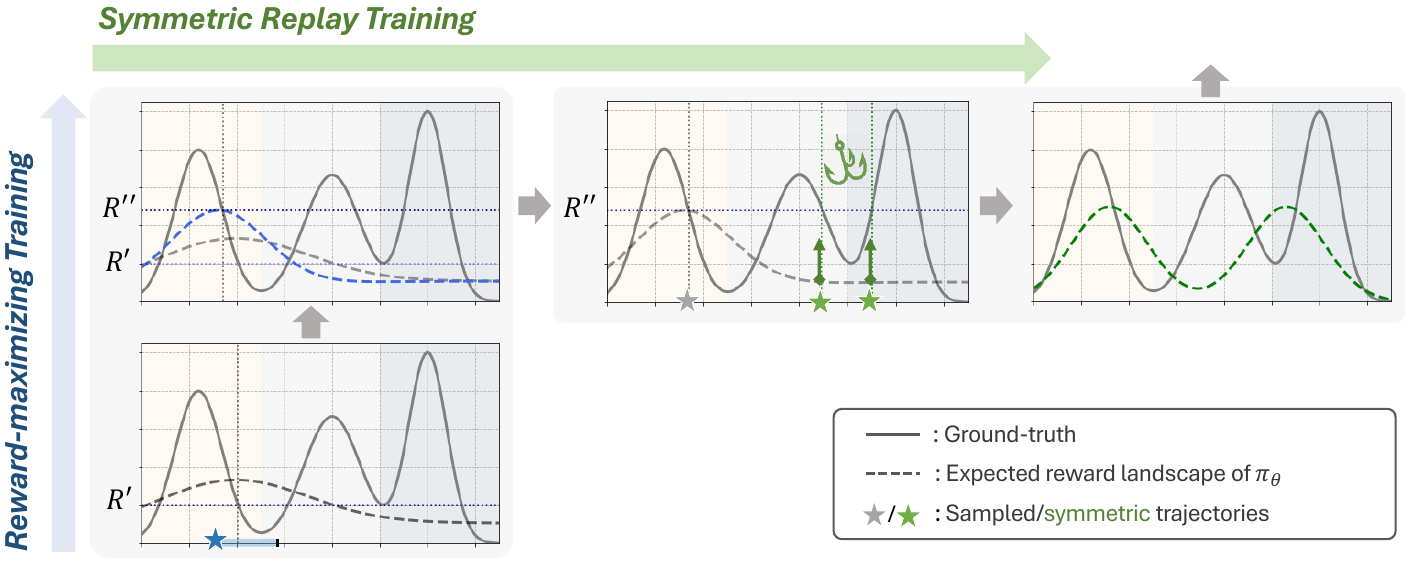}
    \vspace{-5pt}
    \caption{Illustration for two-step training strategies: reward-maximizing training and symmetric replay training.}
    \label{fig:exploration}
    \vspace{-5pt}
\end{figure*}

This paper proposes a simple yet effective add-on method, \textit{symmetric replay training (\ours{})}, which is a generic method that leverages the solution-symmetric nature of combinatorial space.
In DRL for CO, multiple action sequences (i.e., action trajectories) can be mapped to a single combinatorial solution, and the reward is defined on the terminal state (i.e., solution). Previous studies have tried to consider the symmetries in CO, but they utilize numerous samples \citep{kwon2020pomo,kim2022sym} or assume the specific method \citep{yao2022data} or problem \citep{duan2022augment}.
To effectively utilize symmetries in CO, we suggest a decomposed training process in two steps: reward-maximizing training and symmetric replay training. The two-step training also allows efficient exploration in the vast action trajectory space.

Reward-maximizing training is conducted with a conventional DRL algorithm, which involves the exploration of the high-rewarded action trajectories over the entire action space. 
Subsequently, symmetric replay training bootstraps the policy to explore symmetric regions without additional reward evaluation, leading to improved sample efficiency. In detail, the high-rewarded action trajectories obtained in the previous step are symmetrically transformed, and then the policy is trained to imitate these symmetric action trajectories. 
As illustrated in \Cref{fig:exploration}, the symmetric replay training promotes the exploration of under-explored regions containing trajectories equivariant to the high-rewarded trajectories collected from reward-maximizing training. 
% It is noteworthy to emphasize that the symmetric replay does not necessitate additional reward evaluations; \emph{it is free}. 

The two decomposed training steps, reward-maximizing and symmetric replay training, are effectively inter-operated: one seeks the high-reward samples, while the other recycles the explored samples to explore symmetric regions \emph{for free}. Symmetric replay training strategically leverages previously high-rewarded trajectories without necessitating additional reward evaluation. Our policy is trained to maximize the likelihood of high-rewarded trajectories in the symmetric region, encouraging exploration of under-explored areas. 
Additionally, replaying symmetric trajectories incorporates symmetric priors, relieving the overfitting of replay training.
Since the policy perceives these symmetric trajectories as heterogeneous, SRT is advantageous in scaling up the replay loop to further utilize collected samples before re-interacting with the environment.

For a better comprehension of symmetric replay training, this paper explores three straightforward strategies: (1) maximum entropy, (2) adversarial, and (3) importance sampling transformations. They provide theoretical and empirical insights for \ours{}. Note that the maximum entropy transformation is primarily adopted since it is a generic approach with low computational complexity while effectively improving sample efficiency in various tasks.

We empirically demonstrate that symmetric replay training consistently improves the sample efficiency by plugging it into various DRL methods with simple implementation. 
The experimental results show that plugging our method into the state-of-the-art DRL method can achieve superior performance in hardware design optimization and sample-efficient molecular optimization by adding it to the competitive DRL method. Here is a highlight of our contribution:
\begin{enumerate}
\vspace{-5pt}
    \item We propose a generic method, symmetric replay training (SRT), which enhances the sample efficiency of various DRL methods for CO. The proposed method is empirically demonstrated as simple yet effective, by consistently improving various DRL methods, both on-policy and off-policy, in various tasks.
\vspace{-2pt}
    \item We investigate various symmetric transformation strategies for replay training, offering both theoretical and empirical perspectives that could benefit future efforts in SRT-like replay training approaches.
\vspace{-2pt}
    \item We conduct comprehensive experiments with a broad range of DRL methods and various CO tasks. We benchmark the DRL methods from on-policy to off-policy methods across various tasks from classical TSP to practical hardware design optimization, which can contribute to DRL for CO communities.
\vspace{-3pt}
\end{enumerate}

% \clearpage
\section{Related Works} \label{sec:related}

\subsection{Symmetries in DRL for CO}

Several works have been studied to utilize the symmetric nature of combinatorial optimization. Representatively, the Policy Optimization for Multiple Optima \citep[POMO; ][]{kwon2020pomo} and Symmetric Neural Combinatorial Optimization \citep[Sym-NCO; ][]{kim2022sym} are proposed. 
These methods sample multiple trajectories from a single problem by forcing $N$ heterogeneous starting point based on the TSP's cyclic symmetries \citep{kwon2020pomo} or giving symmetric input noise to make wide exploration \cite{kim2022sym}.
Importantly, they use symmetries to effectively evaluate the REINFORCE baseline, which stabilizes training by effectively reducing the variance. 
However, they associate extensive reward evaluation, while ours does not require additional sample evaluations to utilize symmetries.
Furthermore, the results in \cref{append:exp_general_co} demonstrate that SRT can further improve the sample efficiency of these methods. The improvements come from the difference that they take account of symmetries in input space, while ours utilize symmetries in output (action trajectories) space.

Separately, Generative Flow Networks \citep[GFlowNets; ][]{bengio2021gflownet}, which represents the combinatorial space in CO problems with a directed acyclic graph, were proposed. We discuss the relationship with GFlowNet in \cref{sec:gfn}.

\subsection{Data Augmentation}
The symmetric transformation shares a similar concept with data augmentation techniques, such as cropping \citep{laskin2020reinforcement, yarats2021image}, noise injection \citep{laskin2020reinforcement,fan2021secant}, and Mixup \citep{fan2021secant,zhang2017mixup}, all of which can contribute to improved sample efficiency. However, most data augmentation methods focus on continuous space.
Notably, recent efforts have been made to adapt data augmentation NLP tasks \citep{shen2020simple,ren2021text,feng2021survey,kim2021dtd}.
One of the key ideas is to obtain informative samples while keeping the semantics of original sentences, \cite{wang2018switchout,garg2020bae}. 
Notably, \citet{kim2021dtd} introduces a learnable augmentation policy that constructs difficult but not too different samples (DND). 
As our symmetric transformation preserves the solutions by utilizing the nature of CO tasks, augmenting trajectories does not lose semantics. 
Note that the adversarial transformation is designed to find difficult samples akin to DND.

In CO, \citet{duan2022augment} presented a label-preserving augmentation in contrastive learning for the Boolean satisfiability problem. Additionally, \citet{yao2022data} utilized augmentation, including flipping, in supervised learning for the TSP. Lastly, \citet{pmlr-v202-kim23h} suggested an additional loss term using symmetrically augmented data in hardware design.
These works suggested data augmentation based on symmetries, which is integrated into a specific learning framework.
On the other hand, our paper introduces a generic method to enhance diverse DRL methods across various domains.
We empirically show that the separated MLE training of \ours{} is also beneficial compared to data augmentation training; see \cref{append:aug}. Note that SRT uses imitation loss, which allows training with augmented trajectories without introducing importance weights, even in on-policy DRL.

\subsection{Replay Ratio Scaling}
Increasing the number of replay loops is highly related to scaling up the replay ratio, which means the number of parameter updates per environment interaction \citep{wang2016sample, fedus2020revisiting, d2022breaking}. Replay ratio is also known as update-to-data (UTD) ratio \citep{chen2019learning, smith2022walk}. Although the benefits of replay ratio scaling are limited, it has demonstrated improved performance, particularly on well-tuned algorithms \citep{kielak2019recent, chen2019learning, smith2022walk}. Recently, \citet{d2022breaking} achieved better replay ratio scaling with parameter reset strategy by mitigating the loss of ability to generalize on model-free RL.
Most replay ratio scaling approaches assume the utilization of replay training, thus off-policy training, and propose how to further increase replay ratio by preventing overfitting or primacy biases \cite{d2022breaking,nikishin22primacy}. On the other hand, we propose a new replay method both for off-policy and on-policy training in CO. Though SRT demonstrates that it can increase replay ratio less suffering from overfitting, other replay ratio scaling approaches like periodic resetting policy parameters or sophisticated design for experience buffer can be incorporated to enhance sample efficiency further.

We provide further related works in \cref{append:related}.

\section{Background} \label{sec:pre}
Our method aims to improve the sample efficiency of deep reinforcement learning for solving combinatorial optimization with expensive reward function.
Specifically, we consider combinatorial optimization as maximization of the black-box function $f(x)$ over a discrete set $\mathcal{X}$, i.e., 
\begin{equation*}
    \max_{x \in \mathcal{X}}f(x).
\end{equation*}
To solve this problem, we formulate the construction and evaluation of the solution as a Markov decision process (MDP).
In the MDP, we let each state $s$ describe a subsequence of the action trajectory with problem context $\bm{c}$, i.e., $s_t = \{(a_1, \ldots, a_{t-1}), \bm{c}\}$.
The initial state corresponds to an empty, i.e., $s_1=\{ \emptyset, \bm{c}\}$, and the final state corresponds to a complete sequence of actions, i.e., $s_{T}= \{\vec{a}, \bm{c}\}$, giving a solution $x$.
Then, a policy $\pi(s'|s )$ decides a transition between states, $s \rightarrow s'$, by selecting an action $a$ to be updated to the incomplete solution described by the state $s$.
We assume that the transition is deterministic, meaning that the next state is determined by a specific transition function.

We further impose two conditions on the MDP that exploit the prior knowledge about combinatorial optimization problems considered in this work. First, we assume the reward is episodic, i.e., given a terminated action-state trajectory associated with a solution $x$, the reward $R(s_{T}) = f(x)$ and $R(s_{t})=0$ for $t < T$. Note that the $s_{T}$ contains the action sequence $\vec{a}$, giving a solution $x$. 
The next condition is that the action space $\mathcal{A}_{t}$ at each state $s_{t}$ only consists of actions that generate a valid solution. 
% The next condition is about how the action space $\mathcal{A}_{t}$ for action $a_{t}$ made at each state $s_{t}$ only consists of actions that generate a valid solution for the combinatorial optimization.

\section{Methodology} \label{sec:method}

% \subsection{Overview} 

% Our method, \textit{symmetric replay training (\ours{})}, improves the sample efficiency of 
Our method, \textit{symmetric replay training (\ours{})}, is an add-on method that improves the sample efficiency of DRL for CO by replaying the symmetrically transformed solution.
The key idea is to utilize the symmetric natures of CO to generate different trajectories, which induce the same solutions, i.e., solution-preserving transformation; no additional reward evaluations are required as the solution remains unchanged.
\ours{} enhances sample efficiency in two ways: (1) by recycling once-evaluated samples via replay training and (2) by stimulating the exploration of under-explored regions via symmetric trajectories.

Our method repeats the following two policy update loops:
\begin{description}
    \item[Step A:] \textbf{Reward-maximizing training.} Train the (factorized) policy using a conventional episodic reinforcement learning algorithm.
    \item[Step B:] \textbf{Symmetric replay training} 
    \begin{enumerate}
        \item Collect high-rewarded trajectories from Step A.
        \item Randomly sample trajectories using symmetric transformation policy.
        \item Train the policy by imitating the symmetric trajectories.
    \end{enumerate}
\end{description}

Intuitively, our \textbf{Step A} is designed to encourage the policy to exploit high-reward samples via reinforcement learning. Then, \textbf{Step B} aims to promote exploration of the symmetric space by imitating the collected high-reward samples with symmetric transformation. 
Note that the strategies for collecting high-reward samples are flexibly employed according to tasks or DRL methods, such as greedy rollout, reward-prioritized sampling, or top-$K$ selection.

\subsection{Reward-maximizing Training}
Reward-maximizing training corresponds to the original training of deep reinforcement learning (DRL) methods. For simplicity, this step is illustrated using the example of REINFORCE, a widely used DRL algorithm for CO \citep[e.g., ][]{bello2017neural,kool2018attention,kim2022sym}. 
% For a detailed explanation of other DRL methods like the proximal policy optimization \citep[PPO; ][]{schulman2017proximal}, please refer to \cref{append:implementation}. 
In reward-maximizing training, the policy $\pi_\theta$ is trained for a given trajectory $\tau=(s_1, \ldots, s_T)$ using the loss function defined as follows:
\begin{equation*}
\begin{split}
    \mathcal{L}_{\text{RM}} = (R(x) - b) \log \pi_\theta (\tau), \\
    \mbox{where } \pi_{\theta}(\tau) = \prod_{t=1}^{T-1} \pi_{\theta}(s_{t+1}|s_t).
\end{split}
\end{equation*}
Here, $b$ is a baseline that does not depend on $x$.
It is worth emphasizing that the model and training method in reward-maximizing training are not restricted. Thus, our method allows the application of various DRL methods as a base.
% In the following subsections, we describe the details of a solution-preserving symmetric transformation policy and replay training.

\begin{figure}
  % \vspace{-15pt}
  \centering
    \includegraphics[width=0.72\linewidth]{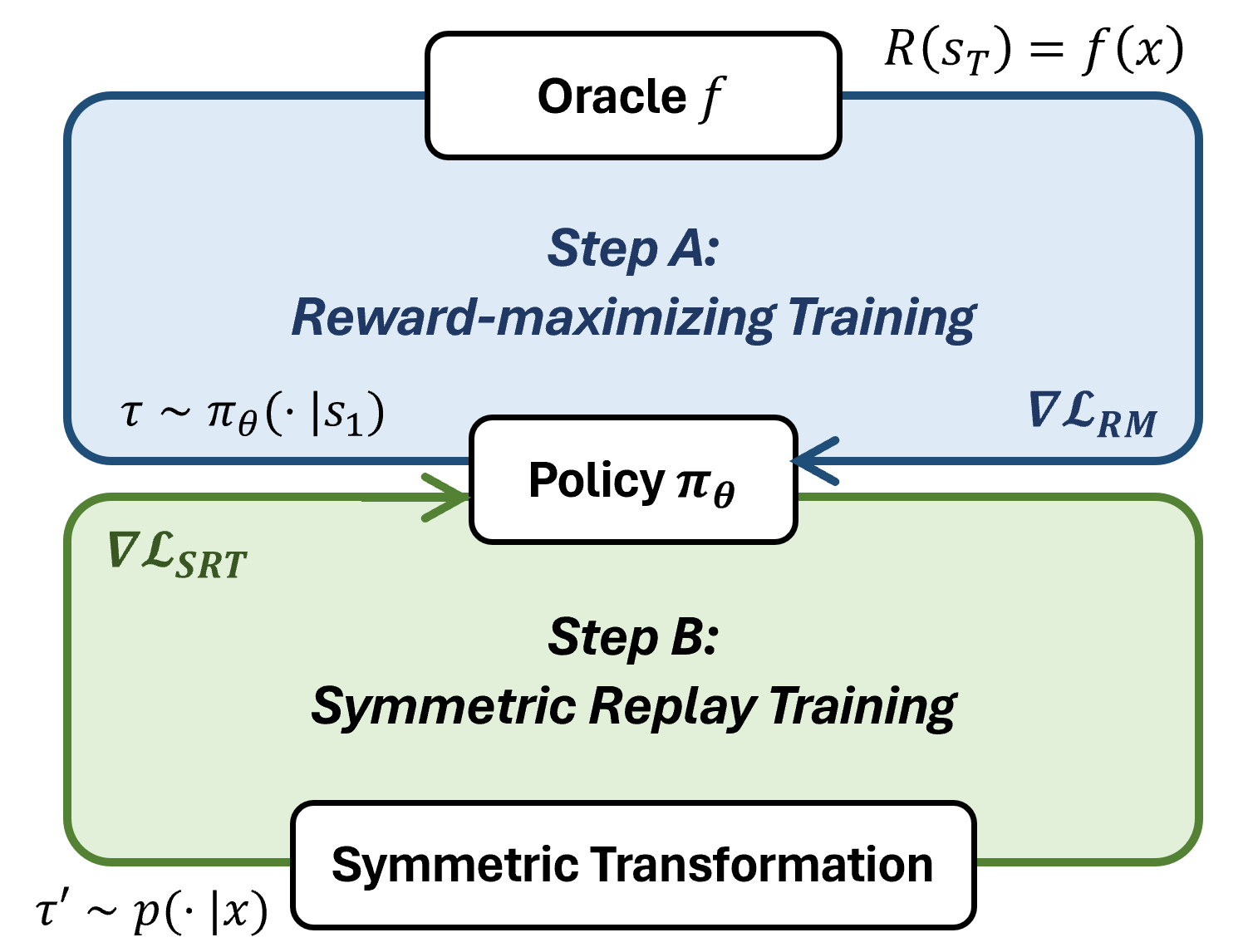}
    \vspace{-5pt}
  \caption{Overview of iterative training} \label{fig:overview}
  \vspace{-5pt}
\end{figure}

\subsection{Symmetric Replay Training}
\subsubsection{Symmetric transformation policy} \label{sec:p_sym}

The symmetric trajectory policy $p(\tau_{\rightarrow x}|x)$ is a probabilistic distribution that samples symmetric trajectories $\tau_{\rightarrow x}$, which have an identical terminal state (i.e., an identical solution $x$). For example, in the traveling salesman problem (TSP), a representative CO problem, shifting the starting points gives the same solution. In the case of TSP with four cities, the sequences 1-2-3-4-1 and 3-4-1-2-3 represent the identical Hamiltonian cycle $x$.

For a better comprehensive understanding of the symmetric transformation policy, we investigate possible choices of  $p(\tau_{\rightarrow x}|x)$: maximum entropy, adversarial, and importance sampling transformations.

\paragraph{Maximum entropy transformation policy.} By setting $p(\tau_{\rightarrow x}|x)$ as a uniform distribution, we can achieve the maximum entropy for the training policy $\pi_\theta(\cdot)$; see \Cref{thm1} in \cref{sec:interplay} for theoretical analysis. Notably, the uniform distribution stands out for its simplicity and $\mathcal{O}(1)$ sampling cost (e.g., in TSP, a symmetric trajectory is determined by simply choosing the starting point and direction), easily applicable to various tasks. 
Therefore, we adopt the maximum entropy policy as a default for our experiments.

\paragraph{Adversarial transformation policy.} An alternative option is to sample a trajectory that has the minimum log-likelihood under $\pi_\theta(\tau)$. Formally, symmetric trajectories are selected as follows:
\begin{equation*}
\min_{\tau \in \mathcal{T}_{\rightarrow x}} \log \pi(\tau),
    % \min_{\theta}\max_{\tau_{\rightarrow x}} \log \pi_\theta(\tau_{\rightarrow x})
\end{equation*}
where $\mathcal{T}_{\rightarrow x}$ denotes a set of all trajectories that give a solution $x$. 
In the case of TSP, where the number of symmetric trajectories is $2N$, where $N$ is the problem size, the sampling cost is $\mathcal{O}(N)$.

\paragraph{Importance sampling transformation policy.}
The last option is to sample trajectories based on importance weights, which are set inversely proportional to the log-likelihood. Formally, symmetric trajectories are sampled from
\begin{equation*}
    % \min_{\theta} \mathbb{E}_{p(\tau|x)} \log \pi_\theta(\tau_{\rightarrow x}), \\
    p(\tau_{\rightarrow x}|x) \propto e^{-\beta\log \pi_\theta(\tau_{\rightarrow x})}.
\end{equation*}
Here, $\beta$ is an inverse temperature, a tunable hyperparameter.
Intuitively, it becomes the maximum entropy transformation as $\beta \rightarrow 0$, and the adversarial transformation as $\beta \rightarrow \infty$; refer to \cref{append:proof2} for formal proof. 
Therefore, having $0 < \beta < \infty$ can be considered a compromised approach between the maximum entropy and adversarial transformations.
Note that evaluating the importance weight requires $\mathcal{O}(N)$ computations in TSP.

\subsubsection{Symmetric replay training} \label{sec:ssd}

The symmetric replay training involves maximum likelihood estimation of $\tau_{\rightarrow x}$ sampled from $p(\tau_{\rightarrow x}|x)$. 
The symmetric replay training loss function is derived as follows:
\begin{equation}\label{eq:ssd_loss}
    \mathcal{L}_{\text{SRT}}(x) = - \mathbb{E}_{p(\tau_{\rightarrow x}|x)}[\log \pi_{\theta}(\tau_{\rightarrow x})]
\end{equation}
The \ours{} loss function is formulated to maximize the log-likelihood of the symmetric trajectories. To adjust the scale of the loss function, we introduce a scaling coefficient. Roughly, $\alpha$ is set to make $\mathcal{L}_{\text{SRT}}$ 10 or 100 times smaller than $\mathcal{L}_{\text{RM}}$. %As a rough guideline, we establish a coefficient that renders the \ours{} loss approximately 10 to 100 times smaller than $\mathcal{L}_{\text{RM}}$ according to the base DRL methods. 
Note that the high-reward samples are obtained by greedy rollout or selecting Top-$K$ samples in the mini-batch, and so on.

Our symmetric replay training significantly benefits from additionally exploring the high-rewarded regions on the symmetric space for free. Moreover, we can explore regions that are likely to have higher rewards but are far from the current regions, since the symmetric action trajectory may have a significant edit distance from the original trajectory.\footnote{\textit{Edit distance} is a measure of similarity between two sequences, defined as the minimum number of operations required to transform one sequence into the other, e.g., \textit{Hamming distance}.}

\subsection{Interplay of Reward-maximizing and Symmetric Replay Training} \label{sec:interplay}
This subsection presents an analysis of the interplay between two iterative steps: reward-maximizing training and symmetric replay training. 
First, we begin with introducing a theorem about the maximization of policy entropy.

\begin{theorem}\label{thm1}

Consider a distribution $\pi_{\theta}(\tau)$ over the trajectory, and $\pi_{\theta}(x)$ over its corresponding solutions.
Let $U(\tau_{\rightarrow x} | x)$ denote an uniform distribution. Then the entropy of $\pi_\theta(\tau)$ can be decomposed and upper bounded as follows: 
\begin{equation}
    \begin{split}
        \mathcal{H}(\pi_{\theta}(\tau)) & \\
        = & \underbrace{\mathcal{H}(\pi_{\theta}(x))}_{\text{Step A}} + \underbrace{\mathbb{E}_{x \sim \pi_{\theta}(x)} \mathcal{H}(p(\tau_{\rightarrow x}|x))}_{\text{Step B}} \\
        \leq & \mathcal{H}(\pi_{\theta}(x) + \mathbb{E}_{x \sim \pi_{\theta}(x)} \mathcal{H}(U(\tau_{\rightarrow x}|x)).
    \end{split}
\end{equation}

\end{theorem}
\emph{Proof.} See \cref{append:proof} for the entire proof. 

In \Cref{thm1}, the policy entropy $\mathcal{H}(\pi_{\theta}(\tau))$ is decomposed into two distinct components: the entropy associated with the solution exploration policy, denoted as $\mathcal{H}(p(x))$, and the entropy for the symmetric transformation policy, expressed as $\mathbb{E}_{x \sim p(x)} \mathcal{H}(p(\tau_{\rightarrow x}|x))$. This decomposition offers an intuitive illustration of our two-step learning process. 
% In Step A, maximizing entropy is related to the policy to explore high-reward solutions. Subsequently, 
In particular, maximizing the second entropy term encourages the search of various symmetric spaces while preserving high-reward solutions. We achieve this by employing uniform symmetric transformation, represented as $p(\tau_{\rightarrow x}|x) = U(\tau_{\rightarrow x}|x)$. This approach enables us to maximize entropy exploration within the symmetric space, facilitating a more comprehensive search of potential solutions.

\subsection{Relationship with GFlowNets} \label{sec:gfn}

Our symmetric transformation policy $p(\tau_{\rightarrow x}|x)$ is similar to GFlowNet's backward policy $P_B(s_t|s_{t+1})$ where both assume the multiple possible (partial) trajectories that leads to identical states: i.e., directed acyclic graph (DAG)-like MDP. While GFlowNets trains policy using a balanced objective with $P_B$, such as trajectory balance \citep{malkin2022trajectory}, which is known to give slow learning signals when trajectory length is large \cite{falet2023delta}, our method makes direct credit assignment into a trajectory drawn from $p(\tau_{\rightarrow x}|x)$, which gives faster learning signals. Accordingly, our method significantly improves GFlowNets training efficiency by using our periodic replay training of maximum likelihood estimation in experiments on three tasks.

% \clearpage
\section{Experiments} \label{sec:exp}
This section presents experiments with three distinct settings: a synthetics scenario and two real-world scenarios of hardware design and sample-efficient molecular optimization. 
For the synthetic setting, we employ traveling salesman problems (TSP), the widely studied CO problems, for precise analyses of the proposed method. We also provide the experiments on other CO problems in \cref{append:exp_general_co}.

In the subsequent sections, we validate the effectiveness of our method in real-world applications. Firstly, we conduct experiments focused on hardware design optimization, particularly addressing the Decap Placement Problem (DPP), a widely recognized problem within the hardware design domain \citep{koo2017fast, kim2021dpp, berto2023rl4co}. Furthermore, we extend our experiments to the Practical Molecular Optimization (PMO) benchmark \citep{gao2022sample}, a well-established benchmark for sample-efficient molecular optimization. All data and codes are available at \href{https://github.com/kaist-silab/symmetric_replay}{https://github.com/kaist-silab/symmetric\_replay}.

\subsection{Synthetic Setting: Traveling Salesman Problems}

% \vspace{-10pt}
% \floatsetup[table]{capposition=top}
\begin{table}[t!]
    \centering
    \caption{Sample efficiency on the synthetic TSP ($N=50$) with four independent seeds. The average costs ($\downarrow$) and the standard deviations are reported.}
\resizebox{\linewidth}{!}
{\begin{tabular}{lccc}
\toprule
Method & $K=100\text{K}$ & $K=2\text{M}$  % & $K=2\text{M}$
\\
\midrule
A2C \citep{bello2017neural}
& 6.780 \footnotesize{$\pm$ 0.208}
% & 6.541 \footnotesize{$\pm$ 0.075}
& 6.129 \footnotesize{$\pm$ 0.021}\\ 
A2C + \ours{} (ours)
& \textbf{6.586 \footnotesize{$\pm$ 0.043}}
% & 6.541 \footnotesize{$\pm$ 0.075}
& \textbf{6.038 \footnotesize{$\pm$ 0.005}} \\ 
\midrule
PG-Rollout \citep{kool2018attention}
& 7.138 \footnotesize{$\pm$ 0.196}
% & 6.708 \footnotesize{$\pm$ 0.077 }
& 6.226 \footnotesize{$\pm$ 0.026} \\
PG-Rollout + \ours{} (ours)
& \textbf{6.879 \footnotesize{$\pm$  0.110}}
% & 6.708 \footnotesize{$\pm$ 0.077 }
& \textbf{6.131 \footnotesize{$\pm$ 0.019}} \\

\midrule

PPO \citep{schulman2017proximal}
& 6.771 \footnotesize{$\pm$ 0.120}
& 6.319 \footnotesize{$\pm$ 0.110} \\
PPO + \ours{} (ours)
& \textbf{6.712 \footnotesize{$\pm$ 0.024}}
& \textbf{6.249 \footnotesize{$\pm$ 0.045}} \\

\midrule
GFlowNet \citep{bengio2021gflownet}
& 6.797 \footnotesize{$\pm$ 0.078}
& 6.195 \footnotesize{$\pm$ 0.010} \\
GFlowNet  + \ours{} (ours)
& \textbf{6.596 \footnotesize{$\pm$ 0.030}}
& \textbf{6.167 \footnotesize{$\pm$ 0.013}} \\

\bottomrule
\end{tabular}
}  
    \label{tab:tsp50}
\end{table}

\begin{figure}
\centering
\includegraphics[width=0.97\linewidth]{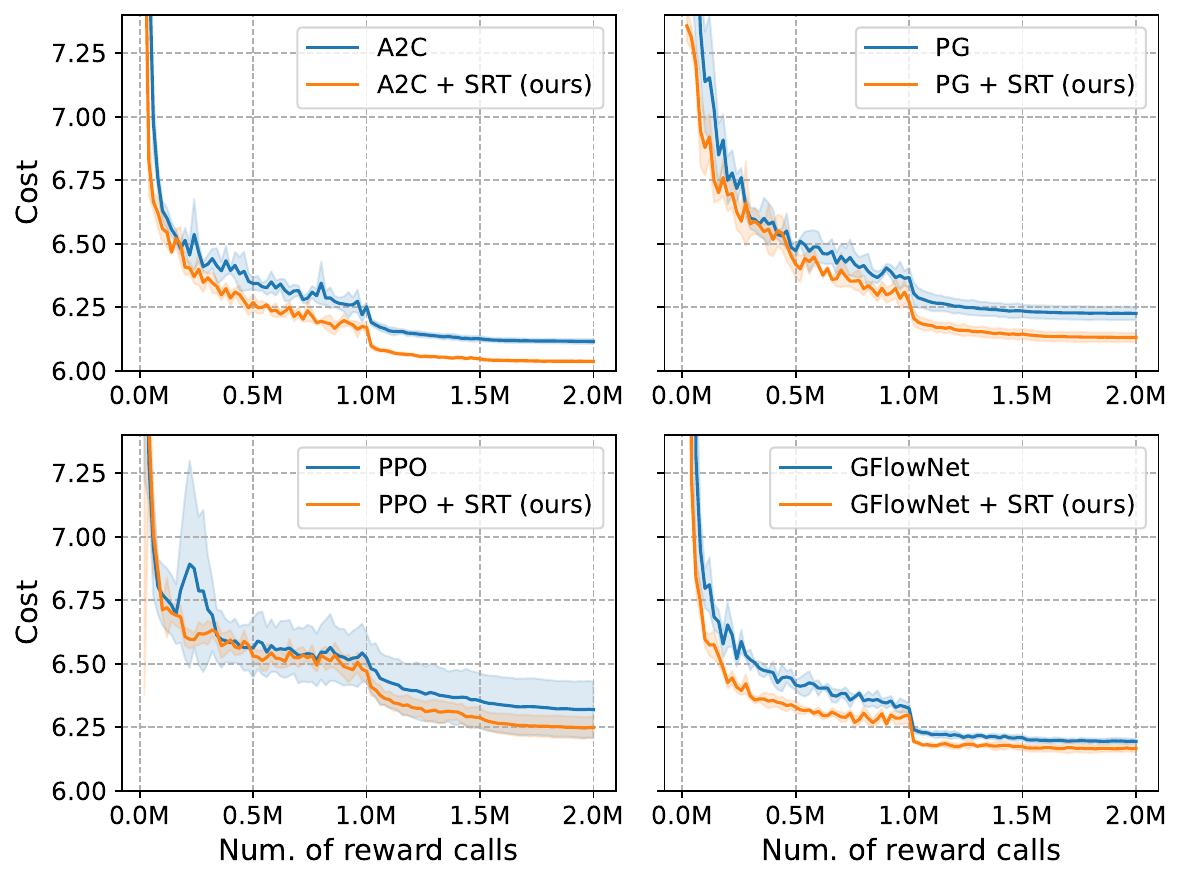}
\vspace{-5pt}
\caption{Optimization curves on TSP ($N=50$)}\label{fig:tsp50}
% \vspace{-5pt}
\end{figure}

In general, CO problems have closed forms of objective functions, which means the reward evaluations are not expensive. 
However, we conduct experiments on the TSP synthetic dataset assuming that the number of computing objective function values is limited; this allows more controlled experiments and more precise analysis.

% \subsubsection{Experimental setting}

\paragraph{Tasks.} Traveling salesman problems (TSP) aim to minimize the distance of a tour that visits all customers and returns to the starting point. In TSP, the distance between consequent customers is defined as an Euclidean distance.
The auto-regressive policy starts from an empty tour and constructs the (partial) tour by iteratively selecting the next visit. 
In TSP, a solution denotes a cycle (i.e., a route) without a designated starting point. Thus, symmetric trajectories are obtained by cyclically shifting $k$ positions to the left or right. Furthermore, in TSP with Euclidean distance, the reversed order of visiting sequence also gives a symmetric action trajectory. We set the maximum reward calls as 2M.

\paragraph{Experimental setting.} 
Our experiments are conducted with AM architecture from \citet{kool2018attention}. We employ various RL methods, including policy gradient with actor-critic \citep{bello2017neural}, policy gradient with greedy rollout \citep{kool2018attention}, Proximal Policy Optimization \citep{schulman2017proximal}, and a Generative Flow Network \citep{bengio2021gflownet}, and additionally implement \ours{} on top of the DRL methods to enhance sample efficiency.
Note that GFlowNet is an off-policy method. %\todo{emphasize various setting.}
We basically follow the hyperparameter configuration used in AM. For additional parameters of PPO and GFlowNet, we systematically evaluate several combinations to identify the most optimal configuration. The details about implementations are provided in \cref{append:implementation}.
We measure the average costs on the validation dataset over the number of reward calls ($K$) in training with four independent random seeds. In symmetric replay training, we gather confidence trajectories from the up-to-date policy via greedy rollout, assuming that the greedy solution gives a relatively high reward trajectory.

% \subsubsection{Results}
\paragraph{Does \ours{} improve sample efficiency?} 
As illustrated in \Cref{fig:tsp50} and detailed in \cref{tab:tsp50}, our approach consistently demonstrates enhanced sample efficiency across various DRL methods. Notably, A2C outperforms the other methods under conditions where the number of available training samples is limited, thereby resulting in the best performance when combined with \ours{}. 
The most substantial improvement facilitated by \ours{} is observed in the case of GFlowNet, where a cost reduction of 3.76\% is achieved when the sample size $K=100\text{K}$. 
In PG-Rollout, a 1.53\% cost reduction is realized with \ours{} when $K=2\text{M}$.

\begin{figure}
  % \vspace{-5pt}
  \centering
    \includegraphics[width=0.62\linewidth]{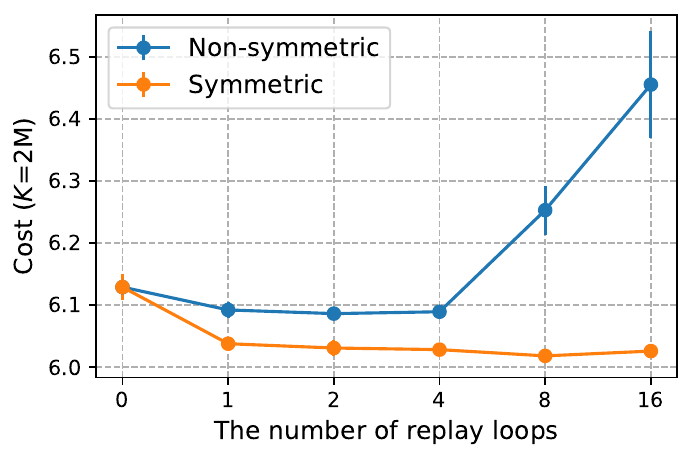}
    \vspace{-5pt}
  \caption{Increasing replay loop} \label{fig:utd}
  \vspace{-5pt}
\end{figure}

\paragraph{Overfitting in replay training.} 
In this section, we investigate the extent to which symmetric replay training can increase the number of replay loops without encountering issues of overfitting. 
To assess the effectiveness of symmetric transformation in replay training, we perform the same experiments without symmetric transformation (the results are denoted as `Non-symmetric' in \Cref{fig:utd}). As illustrated in \Cref{fig:utd}, our symmetric replay training successfully enhances sample efficiency up to replay loops 16. In contrast, non-symmetric replay training experiences diminished performance when reaching when the loop exceeds 8. This suggests that the symmetric transformation provides trajectories that induce the same solution but are distinctive from the policy's perspective, contributing to mitigating overfitting from replaying the restricted set of repetitive trajectories.

\begin{figure}
    \centering
    \includegraphics[width=0.49\linewidth]{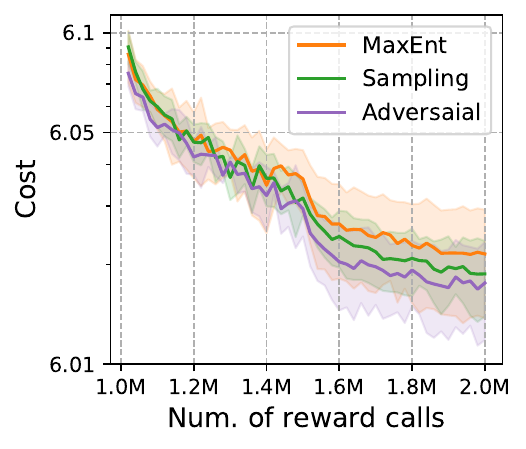}
    \includegraphics[width=0.48\linewidth]{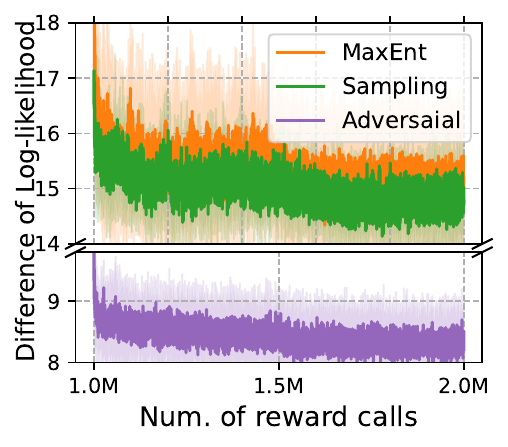}
    \vspace{-5pt}
    \caption{The optimization curve (\textit{left}) and difference between symmetric trajectories (\textit{right}) with different transformation polices.}
    \label{fig:trans}
    \vspace{-5pt}
    % \resizebox{0.5\linewidth}{!}{\input{tables/diff_trans}}
\end{figure}

\paragraph{Comparing different strategies of $p(\tau_{\rightarrow x}|x)$.} 
We conduct experiments to compare different symmetric transformation policies. 
The difference of loglikelihood between symmetric trajectories is measured as follows:
\begin{equation*}
    \frac{1}{|\mathcal{T}_{\rightarrow x}|} \sum_{\tau \in \mathcal{T}_{\rightarrow x}} \log \pi_\theta (\tau) - \min_{\tau \in \mathcal{T}_{\rightarrow x}} \log \pi_\theta (\tau)
\end{equation*}
% where $\mathcal{T}_{\rightarrow x}$ denotes a set of symmetric trajectories giving $x$.
As illustrated in \cref{fig:trans}, the adversarial transformation effectively minimizes log-likelihood differences among symmetric trajectories, thereby enhancing sample efficiency. Despite its effectiveness, the adversarial transformation might not be unavailable when the analytical knowledge of symmetric trajectories is not accessible. On the other hand, the maximum entropy transformation is a more generic approach, so it is advantageous for practical scenarios.

\paragraph{Additional experiments.} We provide the experimental results to verify the effectiveness of each choice by ablating the components. Moreover, we compare \ours{} with experience replay. Ours shows superior performance by not introducing importance weight, leading to low variance and allowing replay training of on-policy methods like A2C. All results are provided in \cref{apped:additional_exp}.

\subsection{Hardware Design Optimization}
\label{sec:dpp_exp}

\paragraph{Tasks.} We employ decoupling capacitor placement problems (DPP), which constitute a fundamental optimization challenge in hardware design. In the context of hardware devices like CPUs and GPUs, a decoupling capacitor (decap) is a critical component responsible for reducing power noises along the power distribution network (PDN). The primary objective of DPP is to identify the optimal arrangement for placing these decaps to maximize the power integrity (PI) objective, which involves computationally expensive evaluations. We tackle two distinct DPP tasks: the chip-package PDN \citep{koo2017fast} and the High Bandwidth Memory (HBM) PDN \citep{jun2017hbm} with 15K limited reward calls. These two tasks exhibit variations in their PI landscapes, presenting unique challenges. In DPP, permuting the decision orders of decap yields symmetric trajectories.

\begin{table}
   \caption{Experimental results of DPP on two different PDN environments. The average rewards ($\uparrow$) and standard deviations are reported with four independent runs.}
    \centering
\resizebox{\linewidth}{!}{

\begin{tabular}{lcc}
\toprule
Method & \makecell{Chip-package\\PDN} & HBM PDN \\
\midrule
A2C \citep{kool2018attention}
& \phantom{1}9.772 \footnotesize{$\pm$ 0.823}
& 25.945 \footnotesize{$\pm$ 0.177} \\ 
A2C + \ours{} (ours)
& \textbf{12.757 \footnotesize{$\pm$ 0.267}}
& \textbf{26.449 \footnotesize{$\pm$ 0.094}} \\ 
\midrule
PG-Rollout \citep{kool2018attention}
& 10.240 \footnotesize{$\pm$ 0.955}
& 25.714 \footnotesize{$\pm$ 0.122} \\
PG-Rollout + \ours{} (ours)
& \textbf{12.601 \footnotesize{$\pm$ 0.467}}
& \textbf{26.355 \footnotesize{$\pm$ 0.013}} \\

\midrule

PPO \citep{schulman2017proximal}
& \phantom{1}9.821 \footnotesize{$\pm$ 0.411}
& 25.907 \footnotesize{$\pm$ 0.068} \\
PPO + \ours{} (ours)
& \textbf{11.279 \footnotesize{$\pm$ 0.511}} 
& \textbf{26.322 \footnotesize{$\pm$ 0.141}}  \\

\midrule
GFlowNet \citep{bengio2021gflownet}
& 9.740 \footnotesize{$\pm$ 0.245}
& 25.845 \footnotesize{$\pm$ 0.103} \\
GFlowNet + \ours{} (ours)
& \textbf{12.784 \footnotesize{$\pm$ 0.186}}
& \textbf{26.469 \footnotesize{$\pm$ 0.057}} \\

\bottomrule
\end{tabular}}
\label{tab:dpp_rst}
% \vspace{-10pt}
\end{table}

\begin{figure}
\centering
    \includegraphics[width=0.97\linewidth]{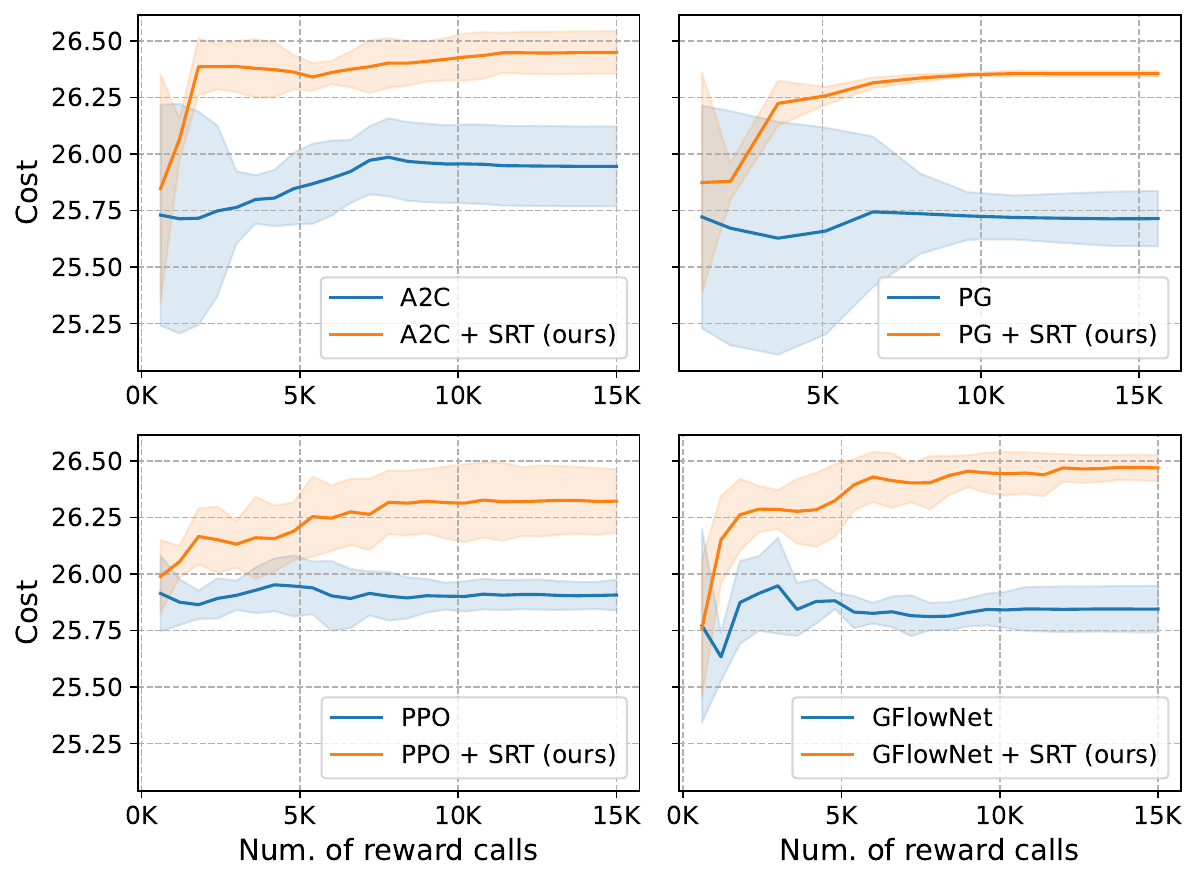}
    \vspace{-5pt}
\caption{Optimization curves on HBM PDN}\label{fig:dpp_rst}
\vspace{-5pt}
\end{figure}

\paragraph{Experimental setting.} We employ a Device Transformer \citep[DevFormer; ][]{pmlr-v202-kim23h} as our base neural architecture, originally designed for offline DPP tasks.
In an online optimization context involving interactions with the DPP environment, we integrate well-established DRL methods (A2C, PG, PPO, and GFlowNet) on DevFormer to maximize effectiveness. Similar to the experiment on TSP, we adjust hyperparameters by examining several combinations since DPP tasks have different reward scales. The details are provided in \cref{append:implementation}. 
Lastly, we use higher-than-average samples in the online batch as high-rewarded samples.
To verify the effectiveness of our methods, \ours{} is built on these DRL methods to enhance sample efficiency and compare the resulting rewards.

\paragraph{Results.} 
Addressing DPP tasks proved challenging due to the limited online reward calls available. For instance, the PG-Rollout method exhibited high variances, making it less effective at exploring the DPP solution space. This challenge stems from the inherent symmetry of DPP, where multiple trajectories could lead to identical solutions. Conversely, GFlowNet models, incorporating structured bias to handle solution symmetries, showed improved performance compared to non-symmetric DRL methods like PPO. 
As demonstrated in \cref{tab:dpp_rst} and \Cref{fig:dpp_rst}, significant improvements are observed by applying \ours{} on top of these methods. The most considerable reward improvement is observed in A2C for chip-package PDN, at 30.54\%.
In the case of GFlowNet with \ours{}, exploring focused on the symmetric variants within the high-reward region enhances the sample efficiency by mitigating its underfitting. 
Notably, the wall-clock comparison results in \cref{append:hardward_additional} show that SRT effectively enhances performance by achieving significantly lower costs while maintaining similar or slightly increased runtime.

\subsection{Practical Molecular Optimization Benchmark}
\label{sec:molopt_exp}

\paragraph{Tasks.} We employ practical molecular optimization \citep[PMO; ][]{gao2022sample}, whose reward evaluations are limited up to 10K, and the goal is achieving the highest score within the limited reward calls.
PMO contains 23 tasks based on different score functions called Oracles; a task is a CO problem that maximizes the given score function, such as \texttt{QED} \citep{qed}, \texttt{DRD2} \citep{olivecrona2017molecular} and \texttt{JNK3} \citep{jnk}. For example, \texttt{QED} measures drug safety, while the others measure bioactivities against their corresponding disease targets. 
In de novo molecular optimization, molecules are represented as graphs or strings,\footnote{In this study, we mainly employ SELFreferencIng Embedded Strings \citep[SELFIES; ][]{krenn2020self} for the string-based representation to ensure compliance with chemical constraints.} which have multiple ways for a single molecule.

% \vspace{-15pt}
\begin{table}
\centering
\caption{Experimental results on sample efficient molecular optimization. AUC top-10 ($\uparrow$) is reported with five independent runs. We use oracle ID instead of a name for better readability.\label{tb:main_molopt}}
\resizebox{\linewidth}{!}{
\begin{tabular}{l|cc|cc}
\toprule
Oracle & GFlowNet & \makecell{GFlowNet\\+ \ours{} (ours)} & REINVENT & \makecell{REINVENT\\+ \ours{} (ours)} \\
\midrule
\#1  & 0.459 \footnotesize{$\pm$ 0.028} & \textbf{0.526 \footnotesize{$\pm$ 0.022}} & 0.847 \footnotesize{$\pm$ 0.021} & \textbf{0.889 \footnotesize{$\pm$ 0.008}} \\
\#2  & 0.437 \footnotesize{$\pm$ 0.007} & \textbf{0.448 \footnotesize{$\pm$ 0.010}} & 0.605 \footnotesize{$\pm$ 0.017} & 0.616 \footnotesize{$\pm$ 0.018} \\
\#3  & 0.326 \footnotesize{$\pm$ 0.008} & \textbf{0.345 \footnotesize{$\pm$ 0.011}} & \textbf{0.603 \footnotesize{$\pm$ 0.084}} & 0.590 \footnotesize{$\pm$ 0.040} \\
\#4  & \textbf{0.587 \footnotesize{$\pm$ 0.002}} & 0.582 \footnotesize{$\pm$ 0.004} & 0.630 \footnotesize{$\pm$ 0.013} & \textbf{0.650 \footnotesize{$\pm$ 0.045}} \\
\#5  & 0.601 \footnotesize{$\pm$ 0.055} & \textbf{0.796 \footnotesize{$\pm$ 0.054}} & 0.953 \footnotesize{$\pm$ 0.006} & \textbf{0.960 \footnotesize{$\pm$ 0.005}} \\
\#6  & \textbf{0.700 \footnotesize{$\pm$ 0.005}} & 0.688 \footnotesize{$\pm$ 0.006} & 0.735 \footnotesize{$\pm$ 0.004} & \textbf{0.751 \footnotesize{$\pm$ 0.009}} \\
\#7  & \textbf{0.666 \footnotesize{$\pm$ 0.006}} & 0.657 \footnotesize{$\pm$ 0.010} & 0.800 \footnotesize{$\pm$ 0.016} & \textbf{0.828 \footnotesize{$\pm$ 0.034}} \\
\#8  & 0.468 \footnotesize{$\pm$ 0.211} & \textbf{0.928 \footnotesize{$\pm$ 0.006}} & \textbf{0.940 \footnotesize{$\pm$ 0.014}} & 0.924 \footnotesize{$\pm$ 0.036} \\
\#9  & 0.199 \footnotesize{$\pm$ 0.199} & \textbf{0.628 \footnotesize{$\pm$ 0.024}} & 0.842 \footnotesize{$\pm$ 0.018} & \textbf{0.858 \footnotesize{$\pm$ 0.016}} \\
\#10 & 0.442 \footnotesize{$\pm$ 0.017} & \textbf{0.505 \footnotesize{$\pm$ 0.040}} & 0.574 \footnotesize{$\pm$ 0.101} & \textbf{0.665 \footnotesize{$\pm$ 0.113}} \\
\#11 & 0.207 \footnotesize{$\pm$ 0.003} & \textbf{0.211 \footnotesize{$\pm$ 0.002}} & 0.350 \footnotesize{$\pm$ 0.012} & \textbf{0.351 \footnotesize{$\pm$ 0.015}} \\
\#12 & 0.181 \footnotesize{$\pm$ 0.002} & 0.181 \footnotesize{$\pm$ 0.002} & 0.255 \footnotesize{$\pm$ 0.014} & \textbf{0.260 \footnotesize{$\pm$ 0.010}} \\
\#13 & 0.332 \footnotesize{$\pm$ 0.012} & \textbf{0.339 \footnotesize{$\pm$ 0.004}} & 0.618 \footnotesize{$\pm$ 0.020} & \textbf{0.637 \footnotesize{$\pm$ 0.065}} \\
\#14 & \textbf{0.785 \footnotesize{$\pm$ 0.003}} & 0.784 \footnotesize{$\pm$ 0.003} & \textbf{0.821 \footnotesize{$\pm$ 0.007}} & 0.820 \footnotesize{$\pm$ 0.005} \\
\#15 & \textbf{0.434 \footnotesize{$\pm$ 0.006}} & 0.429 \footnotesize{$\pm$ 0.010} & \textbf{0.538 \footnotesize{$\pm$ 0.030}} & 0.536 \footnotesize{$\pm$ 0.015} \\
\#16 & 0.917 \footnotesize{$\pm$ 0.002} & \textbf{0.922 \footnotesize{$\pm$ 0.002}} & 0.940 \footnotesize{$\pm$ 0.001} & \textbf{0.941 \footnotesize{$\pm$ 0.000}} \\
\#17 & \textbf{0.660 \footnotesize{$\pm$ 0.004}} & 0.652 \footnotesize{$\pm$ 0.008} & 0.749 \footnotesize{$\pm$ 0.015} & \textbf{0.779 \footnotesize{$\pm$ 0.019}} \\
\#18 & 0.464 \footnotesize{$\pm$ 0.003} & \textbf{0.466 \footnotesize{$\pm$ 0.003}} & 0.529 \footnotesize{$\pm$ 0.020} & \textbf{0.538 \footnotesize{$\pm$ 0.010}} \\
\#19 & 0.217 \footnotesize{$\pm$ 0.022} & \textbf{0.282 \footnotesize{$\pm$ 0.013}} & 0.511 \footnotesize{$\pm$ 0.029} & \textbf{0.535 \footnotesize{$\pm$ 0.046}} \\
\#20 & \textbf{0.292 \footnotesize{$\pm$ 0.009}} & 0.291 \footnotesize{$\pm$ 0.007} & 0.459 \footnotesize{$\pm$ 0.011} & \textbf{0.536 \footnotesize{$\pm$ 0.035}} \\
\#21 & \textbf{0.190 \footnotesize{$\pm$ 0.002}} & 0.189 \footnotesize{$\pm$ 0.004} & 0.342 \footnotesize{$\pm$ 0.016} & \textbf{0.372 \footnotesize{$\pm$ 0.032}} \\
\#22 & 0.000 \footnotesize{$\pm$ 0.000} & 0.000 \footnotesize{$\pm$ 0.000} & 0.000 \footnotesize{$\pm$ 0.000} & 0.000 \footnotesize{$\pm$ 0.000} \\
\#23 & 0.353 \footnotesize{$\pm$ 0.024} & \textbf{0.398 \footnotesize{$\pm$ 0.010}} & 0.512 \footnotesize{$\pm$ 0.017} & \textbf{0.514 \footnotesize{$\pm$ 0.015}} \\
\midrule
\textbf{Avg.}
& 0.431 & \textbf{0.489} & 0.615 & \textbf{0.632}
\\
\bottomrule
\end{tabular}

}
\vspace{-5pt}
\end{table}

\begin{figure}[t!]
% \vspace{-3pt}
\centering
    \includegraphics[width=0.97\linewidth]{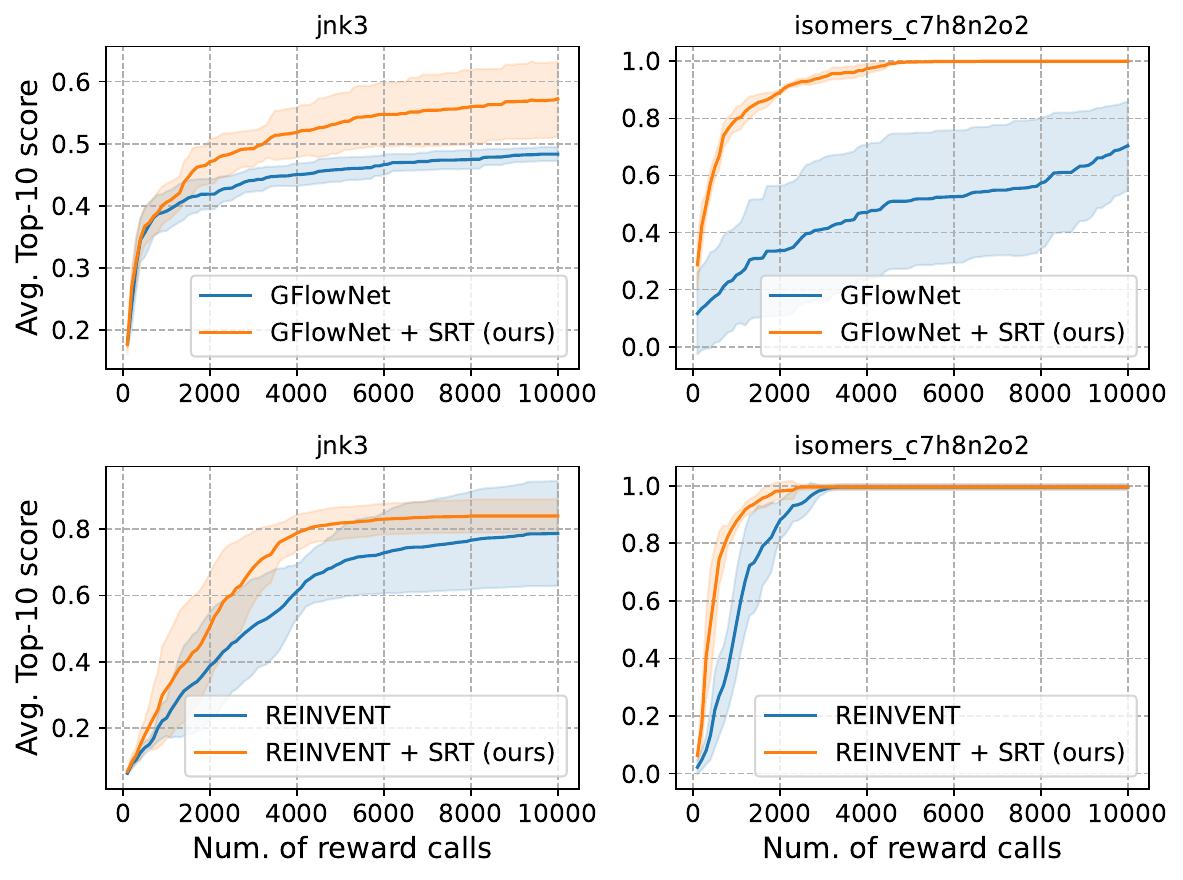}
\vspace{-5pt}
\caption{Average score of Top-$10$ in \texttt{jnk3} (Oracle \#10) and \texttt{isomers\_c7h8n2o2} (Oracle \#8).} \label{fig:molopt_rst}
\vspace{-5pt}
\end{figure}

\paragraph{Experimental setting.} We build \ours{} on the GFlowNet \citep{bengio2021gflownet} and REINVENT \citep{olivecrona2017molecular}. REINVENT is the most competitive DRL method, and GFlowNet is beneficial to generate diverse candidates in the PMO benchmark. 
Note that REINVENT contains the online experience buffer, so we also collect the replay samples from the experience buffer with the same replay size. For GFlowNet, we set the replay size as the same as the sample batch size. 
% \ours{} is compared to not only its base DRL methods, but also MolDQN \citep{zhou2019optimization} and model-based GFlowNet \citep{jain2022biological}. 
% It is noteworthy that REINVENT is considered state-of-the-art (SOTA) in the PMO benchmark.
The performance of the methods is evaluated based on the area under the curve (AUC) to consider a combination of optimization ability and sample efficiency. The AUC of the top 10 average performance is mainly reported since it is essential to find distinct molecular candidates to progress to later stages of development in drug discovery. %Every experiment is conducted with five independent seeds. 
Lastly, as a high-rewarded sample to replay, we gather Top-$K$ samples from the entire pool of generated samples up to that point in training.

\textbf{Results.} As shown in \cref{tb:main_molopt}, \ours{} significantly improves GFlowNet and REINVENT by achieving the enhanced AUC in 12 and 18 oracles out of 23, respectively, not only on average; the statistical analyses are provided in \cref{append:pmo_stat}.
Note that REINVENT is regarded as a state-of-the-art method in PMO, and \ours{} also improves 16 oracles out of 23 in the experiment of REINVENT with SMILES strings; see \cref{append:pmo_additional}. In addition, \Cref{fig:molopt_rst} further demonstrates our method's effectiveness by illustrating the average scores of the top 10 molecules in the \texttt{JNK3} and \texttt{isomer\_c7h8n2o2} tasks, revealing a substantial enhancement in sample efficiency.

\section{Conclusion} \label{sec:conclusion}

This study proposes a new approach, called \textit{symmetric replay training (SRT)}, to enhance the sample efficiency of DRL methods for practical combinatorial optimization problems. 
Our approach improves the sample efficiency by reusing the high-rewarded samples from the policy in the symmetric space, which helps explore new regions without additional reward computation.
Replay training through symmetric transformations enhances the sample efficiency by effectively increasing the replay ratio while mitigating the adverse effects of overfitting.
The proposed method is a generic add-on approach that can synergize with previously proposed constructive DRL methods across various CO problems. 
Since our method is conducted separately (in Step B), we do not introduce additional restrictions on training or alter the architecture of base DRL methods (in Step A). 
Another contribution of our method is the utilization of maximizing log-likelihood loss functions and separated training steps. Therefore, SRT does not introduce importance weight, which often leads to large variances, even when applied with on-policy methods.
The extensive experiments verify that our method consistently enhances sample efficiency in various DRL methods in real-world benchmarks, like hardware design and molecular optimization.

% \clearpage

\section*{Impact Statement}

This research intersects reinforcement learning and combinatorial optimization, areas that inherently involve various industrial and social impacts. However, we believe that this work does not present any unique societal consequences that necessitate specific emphasis beyond those typically associated with these fields.

\section*{Acknowledgments}

The authors are grateful to Haeyeon Kim for the help in implementing the hardware design simulator.
This work was supported by the Institute of Information \& communications Technology Planning \& Evaluation (IITP) grant funded by the Korea government (MSIT) (2022-0-01032, Development of Collective Collaboration Intelligence Framework for Internet of Autonomous Things).

% \section*{Acknowledgements}

% \nocite{lang/ley00}

\bibliography{references}
\bibliographystyle{icml2024}

%%%%%%%%%%%%%%%%%%%%%%%%%%%%%%%%%%%%%%%%%%%%%%%%%%%%%%%%%%%%%%%%%%%%%%%%%%%%%%%
%%%%%%%%%%%%%%%%%%%%%%%%%%%%%%%%%%%%%%%%%%%%%%%%%%%%%%%%%%%%%%%%%%%%%%%%%%%%%%%
% APPENDIX
%%%%%%%%%%%%%%%%%%%%%%%%%%%%%%%%%%%%%%%%%%%%%%%%%%%%%%%%%%%%%%%%%%%%%%%%%%%%%%%
%%%%%%%%%%%%%%%%%%%%%%%%%%%%%%%%%%%%%%%%%%%%%%%%%%%%%%%%%%%%%%%%%%%%%%%%%%%%%%%
\newpage
\appendix
\onecolumn

\section{Proof for Theoretical Analysis in Section 3.2}

\minsu{\subsection{Proof for \cref{thm1}}\label{append:proof}
Consider $\pi_{\theta}(\tau)$ as the distribution over the trajectory. Let $\mathcal{T}_{x}$ denote the space of trajectories associated with the solution $\bm{x}$.
\begin{align*}
    \mathcal{H}(\pi_{\theta}(\tau)) 
    &= -\sum_{\tau \in\mathcal{T}}\pi_{\theta}(\tau) \log \pi_{\theta}(\tau)\\
    &= -\sum_{\bm{x}\in\mathcal{X}}\sum_{\tau_{\rightarrow x} \in \mathcal{T}_{ x}}\pi_{\theta}(\tau) \log \pi_{\theta}(\tau)\\
    &= -\sum_{\bm{x}\in\mathcal{X}}\sum_{\tau_{\rightarrow x} \in \mathcal{T}_{ x}}p(\tau_{\rightarrow x} | \bm{x}) p(\bm{x}) \left(\log p(\tau_{\rightarrow x} | \bm{x}) + \log p(\bm{x}) \right) \\
    &= \mathcal{H}(p(\bm{x})) + \mathbb{E}_{\bm{x} \sim p(\bm{x})} \mathcal{H}(p(\tau_{\rightarrow x}|\bm{x})) \\
    &\leq \mathcal{H}(p(\bm{x})) + \mathbb{E}_{\bm{x} \sim p(\bm{x})} \mathcal{H}(U(\tau_{\rightarrow x}|\bm{x})),
\end{align*}
where $U(\tau |\bm{x})$ is a uniform distribution over action-trajectories associated with the solution $\bm{x}$. The third equality stems from the fact that $\pi_{\theta}(\tau) = \pi_{\theta}(\tau, \bm{x})$ since $\bm{x}$ is fixed given $\tau$. One can show that the final upper-bound is the entropy of distribution obtained from replacing $p(\tau|\bm{x})$ by $U(\tau|\bm{x})$.

\subsection{Extreme Case of Inverse Temperature in Boltzmann Distribution}\label{append:proof2}

\begin{proposition}
Consider $p(\tau)$ to be the Boltzmann distribution of energy $\mathcal{E}(\tau)$ with inverse temperature $\beta$ given by:
\[ p(\tau) \propto e^{-\beta \mathcal{E}(\tau)}. \]
The following two statements are satisfied:

\paragraph{Statement 1}
If $\beta \rightarrow \infty$, then $p(\tau) = \delta_{\tau, \text{argmin} \, \mathcal{E}(\tau)}$, where $\delta_{i,j}$ is the Kronecker delta function.

\paragraph{Statement 2}
If $\beta = 0$, then $p(\tau)$ is a uniform distribution.
\end{proposition}

\begin{proof}[Proof of Statement 1]
Consider the limit as $\beta \rightarrow \infty$. The Boltzmann distribution becomes:
\[ \lim_{\beta \rightarrow \infty} p(\tau) \propto \lim_{\beta \rightarrow \infty}e^{-\beta \mathcal{E}(\tau)}. \]

For any $\tau$ such that $\mathcal{E}(\tau) > \min_{\tau'} \mathcal{E}(\tau')$, the exponential term $e^{-\beta \mathcal{E}(\tau)}$ approaches zero as $\beta$ becomes very large. To see this, let $\epsilon = \mathcal{E}(\tau) - \min_{\tau'} \mathcal{E}(\tau') > 0$. Then, we have:
\[ \lim_{\beta \rightarrow \infty} e^{-\beta \mathcal{E}(\tau)} = \lim_{\beta \rightarrow \infty} e^{-\beta (\min_{\tau'} \mathcal{E}(\tau') + \epsilon)} = 0. \]

Now, for the case where $\tau = \text{argmin} \, \mathcal{E}(\tau)$, the exponential term $e^{-\beta \mathcal{E}(\tau)}$ becomes 1, as $\beta$ does not affect the term when $\tau$ minimizes the energy.

Thus, in the limit $\beta \rightarrow \infty$, we have:
\[ \lim_{\beta \rightarrow \infty} p(\tau) \propto \begin{cases} 1, & \text{if } \tau = \text{argmin} \, \mathcal{E}(\tau), \\ 0, & \text{otherwise.} \end{cases} \]

Now, using the definition of the Kronecker delta function $\delta_{i,j}$:
\[ \lim_{\beta \rightarrow \infty} p(\tau) \propto \delta_{\tau, \text{argmin} \, \mathcal{E}(\tau)}. \]
Thus, Statement 1 holds.
\end{proof}

\begin{proof}[Proof of Statement 2]
For $\beta = 0$, the Boltzmann distribution becomes:
\[ p(\tau) \propto e^{-0 \cdot \mathcal{E}(\tau)} = 1. \]

Therefore, $p(\tau)$ is a constant, and the distribution is uniform, satisfying Statement 2.
\end{proof}
}

\minsu{Let the energy $\mathcal{E}(\tau)$ be a log-likelihood of training policy: $\mathcal{E}(\tau) = \log \pi_{\theta}(\tau)$. Then, statement 1 shows that our important sampling transformation becomes the adversarial transformation. Also, statement 2 shows that our important sampling transformation becomes the maximum entropy (i.e., uniform) transformation.}

\clearpage
\section{Implementation details} \label{append:implementation}

\subsection{Pseudo-code for REINFORCE with \ours}
In this section, we provide a pseudo-code in case of employing REINFORCE as a base DRL method in Step A.

\begin{algorithm}[htb]
   \caption{REINFORCE with \ours{}} \label{alg:srt}
\begin{algorithmic}[1]
    \STATE {\bfseries Input:} objective function $f$, batch size $B$, SRT sample width $L$, a scale coefficient $\alpha$
    \STATE Initialize the number of reward calls $k \gets 0$, 
    \WHILE{$k \leq K$}
    \STATE $s_1^{i} \gets \texttt{GetInitialState()}, \quad \forall i \in \{1, \ldots, B\}$
    \STATE $\tau \sim \pi_\theta(\cdot|s_1)= \prod_{t=1}^{T} \pi_\theta (s_{t+1}|s_t)$
    \STATE \textcolor{purple}{\(\triangleright\) \quad \textsc{Reward Evaluation}}
    \STATE $R(x^i) \gets f(x^i), \quad \forall i \in \{1, \ldots, B\}$
    \STATE $k \gets k + B$ 
    \STATE \textcolor{purple}{\(\triangleright\) \quad \textsc{Step A. Reward-maximizing Training}}
    \STATE $\mathcal{L}_{\text{RM}} \gets \sum_{i=1}^B (R(x^i) - b) \log \pi_\theta(\cdot) $ 
    \STATE $\theta \gets \text{Adam}(\theta, \nabla \mathcal{L}_{\text{RM}})$
    \STATE \textcolor{purple}{\(\triangleright\) \quad \textsc{Step B. Symmetric Replay Training}}
    \STATE $\tau^{i} \gets \texttt{GreedyRollout()}, \quad \forall i \in \{1, \ldots, B\} \qquad$
    \STATE Symmetric transform $\tau$, i.e., $\tau^{i,1}, \ldots \tau^{i,L} \sim p(\tau_{\rightarrow x}|x^i), \quad \forall i \in \{1, \ldots, B\}$
    \STATE $\mathcal{L}_{\text{SRT}} \gets \alpha \sum_{i=1}^B \frac{1}{L} \sum_{l=1}^L   \log \pi_\theta (\tau^{i,l}|s_1^i)$
    \STATE $\theta \gets \text{Adam}(\theta, \nabla \mathcal{L}_{\text{SRT}})$
    \ENDWHILE
\end{algorithmic}
\end{algorithm}

There are various ways to collect high-reward samples in Step B. For example, $\texttt{GreedyRollout}$ can be replaced with other strategies, such as reward-prioritized sampling or collecting Top-$k$ samples.

\subsection{Proximal policy optimization (PPO)}

We use AM architecture \citep{kool2018attention} on TSP and Devformer architecture \citep{pmlr-v202-kim23h} on DPP for parameterizing compositional policy $\pi(\bm{x}|s_1) = \prod_{t=1}^N \pi(a_t|s_t)$. Then, we implement based on the following equation as follows:

\begin{align*}
     \mathcal{L}(\bm{x};s_1) &= \min \left[A(\bm{x};s_1) \frac{\pi(\bm{x}|s_1)}{\pi_{\text{old}}(\bm{x}|s_1)}, A(\bm{x};s_1) \text{clip}\left(\frac{\pi(\bm{x}|s_1)}{\pi_{\text{old}}(\bm{x}|s_1)}, 1-\epsilon, 1+\epsilon\right)\right], \\
    A(\bm{x};s_1) &= R(\bm{x};s_1) - V(s_1),
\end{align*}

where $R$ stands for reward function and $V$ stands for value function. Since we implement PPO on compositional MDP setting, we train value function in the context of $s_1$ by following actor-critic implementation of \citet{kool2018attention}.

\paragraph{Hyperparameters.} We systematically investigate a range of hyperparameter combinations involving different baselines ([rollout, critic]), various values for clipping epsilon ([0.1, 0.2, 0.3]), and numbers of inner loops ([5, 10, 20]). Our observations reveal that the critic baseline consistently enhances training stability across all tasks, leading to reduced variation when modifying the training seeds. The best configurations for each task are provided in \cref{tb:ppo_tuning}.

\begin{table}[ht!]
    \centering
    \caption{Hyperparameter configurations for PPO.}
    % \resizebox{0.7\linewidth}{!}{
    \begin{tabular}{lccc}
    \toprule
        & TSP & Chip-package PDN & HBM PDN \\
        \midrule
        Baseline & critic & critic & critic \\
        Eps. clip & 0.2 & 0.1 &  0.2 \\
        Number of inner loops $k$ & 5 & 20 & 10 \\
    \bottomrule
    \end{tabular}
    \label{tb:ppo_tuning}
\end{table}

\subsection{Generative flow network (GFlowNet)}

Similar to the PPO implementation, we employ the Attention Model (AM) architecture \citep{kool2018attention} on the Traveling Salesman Problem (TSP), and the DevFormer architecture \citep{pmlr-v202-kim23h} on DPP, for parameterizing the compositional forward policy $P_F(\tau|s_1) = \prod_{t=1}^N P_F(a_t|s_t)$. Subsequently, we configure the backward policy $P_B$ as a uniform distribution for all possible parent nodes, following the methodology outlined in \citep{malkin2022trajectory}. Lastly, we parameterize $Z(s_1)$ using a two-layer perceptron with ReLU activation functions, where the number of hidden units matches the embedding dimension of the AM or DevFormer. This two-layer perceptron takes input from the mean of the encoded embedding vector obtained from the encoder of the AM or DevFormer and produces a scalar value to estimate the partition function.

To train the GFlowNet model, we use trajectory balance loss introduced in \citet{malkin2022trajectory} as follows: 

\begin{equation}
    \mathcal{L}(\tau;s_1) = \left(\log \left(\frac{Z(s_1)P_F(\tau|s_1)}{e^{-\beta E(\bm{x};s_1)}P_B(\tau|s_1)}\right)\right)^2 \label{eq:tb}
\end{equation}

The trajectory $\tau$ includes a terminal state represented as $\bm{x}$. Subsequently, we employ an on-policy optimization method to minimize \cref{eq:tb}, with trajectories $\tau$ sampled from the training policy $P_F$. In this context, $E(\bm{x};s_1)$ represents the energy, which is essentially the negative counterpart of the reward $R(\bm{x};s_1)$. 
The hyperparameter $\beta$ plays the role of temperature adjustment in this process.

\paragraph{Hyperparameters.} We explore a spectrum of hyperparameter combinations, varying $\beta$ ([5, 10, 20]) and numbers of inner loops ([2, 5, 10]). The best configurations for each task are provided in \cref{tb:gfn_tuning}.

\begin{table}[ht!]
    \centering
    \caption{Hyperparameter configurations for GFlowNet.}
    \begin{tabular}{lccc}
    \toprule
        & TSP & Chip-package PDN & HBM PDN \\
        \midrule
        % Baseline & rollout & no baseline & no baseline \\
        $\beta$ & 20 & 10 & 10 \\
        Number of inner loops $k$ & 10 & 2 & 2 \\
    \bottomrule
    \end{tabular}
    \label{tb:gfn_tuning}
\end{table}

\clearpage
\section{Experimental details}\label{append:exp}
\subsection{Traveling salesman problems (TSP)}
Since we employ the AM architecture, we use the same hyperparameters for the model architecture and training parameters except for the batch and epoch data sizes.\footnote{AM: \url{https://github.com/wouterkool/attention-learn-to-route}}
Initially, the Attention Model (AM) employed a batch size of 512 and an epoch data size of 1,280,000. Notably, the evaluation of the greedy rollout baseline was conducted every epoch. When the number of available training samples is constrained, utilizing a smaller batch size and epoch data size becomes advantageous. Consequently, we adjusted these parameters to be 100 for batch size and 10,000 for epoch data size.
In symmetric replay training (Step B), the scale coefficients are meticulously set to scale the SRT loss. As a rough guideline, we establish a coefficient that renders the SRT loss approximately 10 to 100 times smaller than the RL loss. 
Additionally, for the number of symmetric transformations ($L$ in \cref{eq:ssd_loss}) is set as the number of inner loops. See \cref{tb:tsp_hyper} in details.

\begin{table}[h!]
    \centering
    \caption{Scale coefficient and the number of symmetric transformations in TSP.}
    % \resizebox{0.5\linewidth}{!}{
    \begin{tabular}{lcccc}
    \toprule
        & A2C & PG-Rollout & PPO & GFlowNet \\
        \midrule
        Scale coefficient & 0.001 & 0.001 & 0.00001 & 0.1 \\
        $L$ & 1 & 1 & 5 ($= k$) & 10 ($= k$) \\
    \bottomrule
    \end{tabular}
    \label{tb:tsp_hyper}
\end{table}

\subsection{Decap placement problems (DPP)}
Similar to the experiments on TSP, we follow the setting of DevFormer.\footnote{DevFormer: \url{https://github.com/kaist-silab/devformer}} We set the batch size as 100 and epoch data size as 600. Note that the maximum number of reward calls is set 15K,, a considerably smaller limit compared to TSP. Regarding the scale coefficient and the number of symmetric transformations, we maintain consistency with the principles applied in the TSP experiments as follows:

\begin{table}[ht!]
    \centering
    \caption{Scale coefficient and the number of symmetric transformations in DPP tasks.}
    \begin{tabular}{lcccc}
    \toprule
        & A2C & PG-Rollout & PPO & GFlowNet \\
        \midrule
        Scale coefficient & 0.01 & 0.01 & 0.01 & 0.1 \\
        $L$ & 1 & 1 & 20 ($= k$) & 2 ($= k$) \\
    \bottomrule
    \end{tabular}
    \label{tb:dpp_hyper}
\end{table}

\subsection{Practical molecular optimization (PMO)}

We basically follow the experimental setting (e.g., batch size) in the practical molecular optimization (PMO) benchmark.\footnote{Practical molecular optimization: \url{https://github.com/wenhao-gao/mol_opt}}
In symmetric replay training, we utilize reward-prioritized sampling for the online buffer, which contains molecules generated during online learning.
For the REINVENT, where the replay buffer is already incorporated, we set the number of replaying samples equal to the replay buffer size, i.e., 24, and the scale coefficient to 0.001. Regarding the GFlowNet method, we configure the number of replaying samples to match the batch size of 64. Furthermore, we set the scale coefficient to 1.0, given that the RL loss in GFlowNet is much higher compared to REINVENT.

\clearpage
\section{Additional Experiments}  \label{apped:additional_exp}
\subsection{Ablation study for the number of symmetric trajectories in SRT}
\begin{figure}[ht]
% \vspace{-10pt}
\centering
    \includegraphics[width=0.4\linewidth]{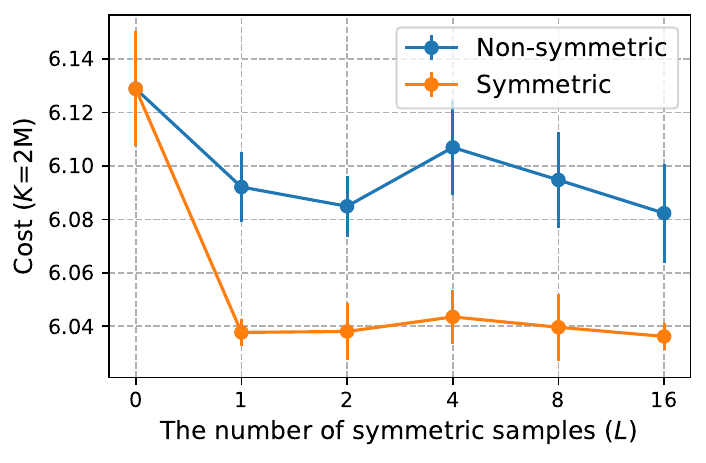}
    % \caption{Increasing sample width $L$}\label{fig:l_abl}
\caption{Ablation study for the sample width in \ours{}} \label{fig:l_abl}
\end{figure}

We conducted the ablation study for varying sample width in replay training. `Non-symmetric' denotes the replay training without symmetric transformation; thus, increased the number of trajectories gives duplicated samples. \Cref{fig:l_abl} shows that the symmetric replay training consistently gives better performance and robust to the choice of $L$.

\subsection{Comparison with Experience Replay on TSP50} 
\label{append:er}

We conduct comparative experiments with experience replay and ours using A2C on TSP50. This requires importance sampling weight as follows:
$$\mathbb{E}_{\tau \sim \pi(\cdot|s_1)}R(\bm{x}) = \mathbb{E}_{\tau \sim q(\cdot|s_1)}\left[ \frac{\pi(\tau|s_1)}{q(\tau|s_1)}R(\bm{x}) \right] = \mathbb{E}_{\tau \sim q(\cdot|s_1)} \left[\prod_{t=1}^{T}\frac{\pi(s_{t+1}|s_t)}{q(s_{t+1}|s_t)}R(x)\right],$$
where $\pi(\cdot|s_1)$ and $q(\cdot|s_1)$ are the training policy and behavior policy, respectively, and $x$ is a solution (i.e., the terminal state).

\begin{figure*}[ht]
\centering
    \includegraphics[width=0.45\textwidth]{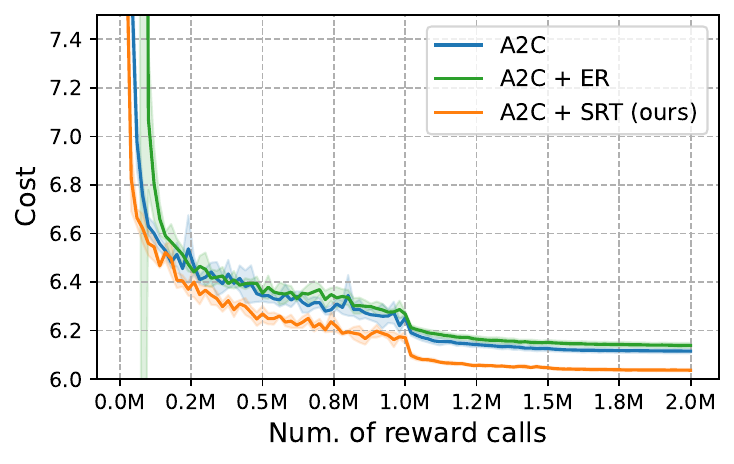}

\caption{Comparative experiments with experience replay and \ours{}. The average validation costs over the computation budget are measured.}\label{fig:er}
\end{figure*}

The results show that the existing experience replay method can suffer from high variance because of importance sampling, leading to the degradation of performance. 
Note that the importance weights are required even without symmetric transformation when using on-policy training, like A2C.
On the other hand, ours does not require the importance sampling weight since SRT uses imitation learning loss.

\subsection{Comparison with Data Augmentation on TSP50}  \label{append:aug}

Without \ours{}, symmetric trajectories can be utilized like data augmentation. In this experiment, we train the policy both using sampled trajectories and symmetric (i.e., augmented) trajectories. 
When on-policy methods like A2C, the importance weights are required for symmetric trajectories similar to the previous section.
The policy is updated only using the reward-maximizing loss $\mathcal{L}_{\text{RM}}(\tau) + \mathcal{L}_{\text{RM}}(\tau'),$ where $\tau' \sim p(\hat{\tau}_{\rightarrow x}|x)$. Here, $\hat{\tau}$ is obtained greedy-rollout; see \cref{alg:srt}.
The experiments are conducted with the A2C method on TSP with $N=50$.

\begin{figure}[ht]
% \vspace{-10pt}
\centering
    \includegraphics[width=0.6\linewidth]{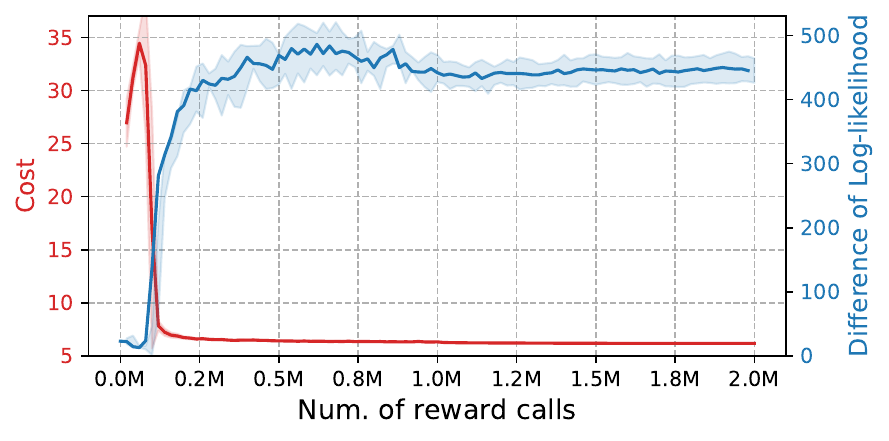}
\caption{Optimization curve (\textit{\textcolor{red}{red}}) and difference of log-likelihood between symmetric trajectories (\textit{\textcolor{cyan}{blue}}) over training steps with reward-maximizing training of symmetric trajectories.}\label{fig:aug}
\end{figure}

\begin{table}[ht]
    \centering
    % \small
    \caption{The average cost and difference of log-likelihood on TSP50 with reward-maximizing training of symmetric trajectories}
    \begin{tabular}{lcc}
    \toprule
        & 100K & 2M \\
    \midrule
        Avg. Cost ($\downarrow$) & 16.851 $\pm$ 11.162 & 6.179 $\pm$ 0.0029 \\
        Log-likelihood Difference ($\downarrow$) & 23.647 $\pm$ 22.331 & 445.044 $\pm$ 19.053 \\
    \bottomrule
    \end{tabular}
    \label{tab:aug}
\end{table}

As depicted in \cref{fig:aug} and \cref{tab:aug}, reward-maximizing training tends to seek one of minimizing cost or reducing differences of symmetric trajectories, not both. 
In the early stage of training, the cost tends to go higher because the sampled and augmented trajectories provide dis-aligned information, as the clearly increased average costs compared to the original A2C method (16.851 $>>$ 6.780). From a certain point, the cost is decreased with a significant increase in the difference of log-likelihood.
This implies that the symmetric nature of CO needs to be utilized carefully so as not to harm the performance of DRL methods, especially when on-policy learning is used.

\clearpage
\subsection{Additional results on hardware design optimization} \label{append:hardward_additional}

To verify the benefit of SRT, we also report the overall runtime (for 15K evaluations) of the hardware design optimization tasks as follows.

\begin{table}[ht]
\centering
\caption{Wall-clock time of whole training in hardware design optimization tasks}
\resizebox{0.525\textwidth}{!}{\begin{tabular}{lrr}
\toprule
 & Chip-package PDN & HBM PDN \\
\midrule
A2C & 4,204.50 ($\pm$342.90) & 33,289.75 ($\pm$682.85) \\
A2C + SRT & 4,265.25 ($\pm$425.38) & 33,294.75 ($\pm$1,722.59) \\
\midrule
PG-Rollout & 3,747.00 ($\pm$516.97) & 34,251.25 ($\pm$573.18) \\
PG-Rollout + SRT & 3,795.25 ($\pm$803.31) & 30,351.25 ($\pm$1,150.44) \\
\midrule
PPO & 3,884.25 ($\pm$81.15) & 32,148.75 ($\pm$623.47) \\
PPO + SRT & 4,700.00 ($\pm$58.26) & 32,479.25 ($\pm$168.71) \\
\midrule
GFlowNet & 3,528.75 ($\pm$26.66) & 35,968.00 ($\pm$3,942.48) \\
GFlowNet + SRT & 3,528.25 ($\pm$17.80) & 35,867.75 ($\pm$5,307.87) \\
\bottomrule
\end{tabular}}
\end{table}

It is worth emphasizing that SRT is beneficial when the reward evaluation costs are more expensive than the additional costs from SRT. Furthermore, SRT allows GPU batch processing, resulting much smaller wall-clock runtime compared to reward evaluation with simulation. The results show that SRT effectively enhances performance by achieving significantly lower costs while maintaining similar or slightly increased runtime, which is negligible compared to the simulation time. We will include this result in the manuscript.

\clearpage
\subsection{Statistical analysis on molecular optimization}
\label{append:pmo_stat}

We perform a t-test to compare the average AUC10 scores, setting the null hypothesis as $\mu_{Base} = \mu_{SRT}$. The resulting p-values are 0.0045 and 0.0015 for REINVENT and GFlowNets, respectively. Thus, the statistical analysis indicates that SRT enhances sample efficiency significantly for both REINVENT and GFlowNets.
\begin{table}[ht]
\centering
\caption{The results of t-statistical test}
\resizebox{0.75\textwidth}{!}{\begin{tabular}{l|rr|rr}
\toprule
 % & \multicolumn{2}{c}{REINVENT} & \multicolumn{2}{c}{GFlowNet} \\
% \cmidrule(r){2-3} \cmidrule(r){4-5}
 & REINVENT & REINVENT + \ours & GFlowNet & GFlowNet + \ours \\
\midrule
Mean & 0.615 & 0.633 & 0.431 & 0.489 \\
Variance &  1.32E-05 & 7.36E-05 & 8.98E-06 & 7.83063E-05 \\
Observations & 5 & 5 & 5 & 5 \\
Hypothesized Mean Diff. & 0 & & 0 & \\
t Stat & -4.1332 & & -6.4356 & \\
P($T \leq t$) one-tail & \textbf{0.0045} & & \textbf{0.0015} \\
t Critical one-tail & 2.0150 & & 2.1318 & \\
\bottomrule
\end{tabular} 
}
\end{table}

\clearpage
\subsection{Additional results on molecular optimization} \label{append:pmo_additional}

In this subsection, we provide the additional results of the black-box optimization method in the PMO benchmark.
Gaussian process Bayesian optimization \citep[GPBO; ][]{tripp2021fresh} and Graph Genetic Algorithm \citep[Graph GA; ][]{jensen2019graph} are included.
Note that GPBO employs Graph GA when optimizing the GP acquisition function.
Though Graph GA demonstrates powerful performance, designing operators, such as crossover and mutation, greatly affects performance \citep{li2022evolutionary} Thus, careful algorithm design, which requires specific domain knowledge, is necessitated whenever there is a change in tasks.
The results show that \ours{} outperforms other black-box optimization methods by improving the on-policy RL method, REINVENT.

Additionally, we apply our method on REINVENT with SMILES, the simplified molecular-input line-entry system \citep{weininger1988smiles}, which is another string-based representation. It is worth mentioning that REINVENT-SMILES gives the best performance in practical molecular optimization benchmark \citep{gao2022sample}. Noticeably, the results in \cref{tb:append_molopt} demonstrate that our achieves improved performances on 16 out of 23 oracles compared to REINVENT-SMILES.

\begin{table}[ht]
\centering
\caption{Experimental results on sample efficient molecular optimization, including REINVENT-SMILES and REINVENT-SMILES with ours. Area under the curve of top-10 average property value ($\uparrow$). Note that the improved cost via SRT is denoted in \textbf{bold}. \label{tb:append_molopt}}
\resizebox{0.8\textwidth}{!}{\begin{tabular}{l|ccc|c}
\toprule
% Oracle & GFlowNet-AL & GPBO & REINVENT & \makecell{REINVENT\\+ \ours{} (ours)} \\
Oracle & GPBO & Graph GA & \makecell{REINVENT\\SMILES} & \makecell{REINVENT-SMILES\\+ \ours{} (ours)} \\
\midrule
albuterol\_similarity 
% & 0.386 \footnotesize{$\pm$ 0.008} 
& 0.896 \footnotesize{$\pm$ 0.009} 
& 0.838 \footnotesize{$\pm$ 0.027} 
& 0.881 \footnotesize{$\pm$ 0.016} 
& \textbf{0.890 \footnotesize{$\pm$ 0.020}} \\
amlodipine\_mpo 
% & 0.432 \footnotesize{$\pm$ 0.001} 
& 0.577 \footnotesize{$\pm$ 0.042} 
& 0.649 \footnotesize{$\pm$ 0.014} 
& 0.645 \footnotesize{$\pm$ 0.018} & \textbf{0.654 \footnotesize{$\pm$ 0.034}} \\
celecoxib\_rediscovery
% & 0.266 \footnotesize{$\pm$ 0.004} 
& 0.733 \footnotesize{$\pm$ 0.026} 
& 0.682 \footnotesize{$\pm$ 0.127} 
& 0.719 \footnotesize{$\pm$ 0.016} & \textbf{0.734 \footnotesize{$\pm$ 0.059}} \\
deco\_hop 
% & 0.583 \footnotesize{$\pm$ 0.001} 
& 0.620 \footnotesize{$\pm$ 0.008} 
& 0.601 \footnotesize{$\pm$ 0.004} 
& 0.665 \footnotesize{$\pm$ 0.042} & 0.643 \footnotesize{$\pm$ 0.009} \\
drd2 
% & 0.443 \footnotesize{$\pm$ 0.047} 
& 0.933 \footnotesize{$\pm$ 0.014} 
& 0.968 \footnotesize{$\pm$ 0.006} 
& 0.957 \footnotesize{$\pm$ 0.007} & \textbf{0.966 \footnotesize{$\pm$ 0.007}} \\
fexofenadine\_mpo 
% & 0.690 \footnotesize{$\pm$ 0.001} 
& 0.723 \footnotesize{$\pm$ 0.002} 
& 0.773 \footnotesize{$\pm$ 0.014} 
& 0.780 \footnotesize{$\pm$ 0.012} & \textbf{0.787 \footnotesize{$\pm$ 0.011}} \\
gsk3b 
% & 0.591 \footnotesize{$\pm$ 0.005} 
& 0.878 \footnotesize{$\pm$ 0.018} 
& 0.792 \footnotesize{$\pm$ 0.092} 
& 0.881 \footnotesize{$\pm$ 0.036} & 0.877 \footnotesize{$\pm$ 0.030} \\
isomers\_c7h8n2o2 
% & 0.788 \footnotesize{$\pm$ 0.023} 
& 0.912 \footnotesize{$\pm$ 0.023} 
& 0.944 \footnotesize{$\pm$ 0.030} 
& 0.941 \footnotesize{$\pm$ 0.012} & \textbf{0.961 \footnotesize{$\pm$ 0.008}} \\
isomers\_c9h10n2o2pf2cl 
% & 0.561 \footnotesize{$\pm$ 0.009} 
& 0.542 \footnotesize{$\pm$ 0.383} 
& 0.831 \footnotesize{$\pm$ 0.018} 
& 0.840 \footnotesize{$\pm$ 0.026} & \textbf{0.874 \footnotesize{$\pm$ 0.021}} \\
jnk3 
% & 0.356 \footnotesize{$\pm$ 0.016} 
& 0.588 \footnotesize{$\pm$ 0.095} 
& 0.677 \footnotesize{$\pm$ 0.120} 
& 0.777 \footnotesize{$\pm$ 0.030} & \textbf{0.813 \footnotesize{$\pm$ 0.070}} \\
median1 
% & 0.191 \footnotesize{$\pm$ 0.003} 
& 0.288 \footnotesize{$\pm$ 0.003} 
& 0.265 \footnotesize{$\pm$ 0.016} 
& 0.372 \footnotesize{$\pm$ 0.014} & 0.364 \footnotesize{$\pm$ 0.002} \\
median2 
% & 0.176 \footnotesize{$\pm$ 0.004} 
& 0.298 \footnotesize{$\pm$ 0.005} 
& 0.268 \footnotesize{$\pm$ 0.013} 
& 0.282 \footnotesize{$\pm$ 0.002} & \textbf{0.289 \footnotesize{$\pm$ 0.012}} \\
mestranol\_similarity 
% & 0.300 \footnotesize{$\pm$ 0.006} 
& 0.659 \footnotesize{$\pm$ 0.108} 
& 0.550 \footnotesize{$\pm$ 0.032} 
& 0.632 \footnotesize{$\pm$ 0.041} & \textbf{0.658 \footnotesize{$\pm$ 0.030}} \\
osimertinib\_mpo 
% & 0.787 \footnotesize{$\pm$ 0.005} 
& 0.788 \footnotesize{$\pm$ 0.003} 
& 0.818 \footnotesize{$\pm$ 0.007} 
& 0.833 \footnotesize{$\pm$ 0.010} & \textbf{0.840 \footnotesize{$\pm$ 0.007}} \\
perindopril\_mpo 
% & 0.418 \footnotesize{$\pm$ 0.005} 
& 0.495 \footnotesize{$\pm$ 0.005} 
& 0.498 \footnotesize{$\pm$ 0.009} 
& 0.536 \footnotesize{$\pm$ 0.015} & \textbf{0.538 \footnotesize{$\pm$ 0.009}} \\
qed 
% & 0.899 \footnotesize{$\pm$ 0.002} 
& 0.936 \footnotesize{$\pm$ 0.001} 
& 0.939 \footnotesize{$\pm$ 0.001} 
& 0.941 \footnotesize{$\pm$ 0.000} & 0.941 \footnotesize{$\pm$ 0.000} \\
ranolazine\_mpo 
% & 0.630 \footnotesize{$\pm$ 0.003} 
& 0.737 \footnotesize{$\pm$ 0.007} 
& 0.716 \footnotesize{$\pm$ 0.011} 
& 0.770 \footnotesize{$\pm$ 0.005} & \textbf{0.796 \footnotesize{$\pm$ 0.008}} \\
scaffold\_hop 
% & 0.459 \footnotesize{$\pm$ 0.002} 
& 0.536 \footnotesize{$\pm$ 0.007} 
& 0.506 \footnotesize{$\pm$ 0.016} 
& 0.552 \footnotesize{$\pm$ 0.023} & \textbf{0.562 \footnotesize{$\pm$ 0.014}} \\
sitagliptin\_mpo 
% & 0.176 \footnotesize{$\pm$ 0.005} 
& 0.422 \footnotesize{$\pm$ 0.008} 
& 0.486 \footnotesize{$\pm$ 0.007} 
& 0.466 \footnotesize{$\pm$ 0.039} & \textbf{0.483 \footnotesize{$\pm$ 0.019}} \\
thiothixene\_rediscovery 
% & 0.264 \footnotesize{$\pm$ 0.008} 
& 0.565 \footnotesize{$\pm$ 0.032} 
& 0.494 \footnotesize{$\pm$ 0.010} 
& 0.544 \footnotesize{$\pm$ 0.029} & 0.539 \footnotesize{$\pm$ 0.027} \\
troglitazone\_rediscovery 
% & 0.184 \footnotesize{$\pm$ 0.002} 
& 0.417 \footnotesize{$\pm$ 0.025} 
& 0.421 \footnotesize{$\pm$ 0.041} 
& 0.456 \footnotesize{$\pm$ 0.021} & 0.428 \footnotesize{$\pm$ 0.026} \\
valsartan\_smarts 
& 0.000 \footnotesize{$\pm$ 0.000} 
& 0.000 \footnotesize{$\pm$ 0.000} 
& 0.182 \footnotesize{$\pm$ 0.363} & 0.000 \footnotesize{$\pm$ 0.000} \\
zaleplon\_mpo 
% & 0.284 \footnotesize{$\pm$ 0.009} 
& 0.456 \footnotesize{$\pm$ 0.019} 
& 0.449 \footnotesize{$\pm$ 0.012} 
& 0.533 \footnotesize{$\pm$ 0.009} & \textbf{0.555 \footnotesize{$\pm$ 0.015}} \\
\midrule
\textbf{Average} & 0.609 & 0.616 & 0.660 & 0.661 \\
% \textbf{Average} & 0.429 & 0.609 & 0.613 & 0.633 \\
% \textbf{Num. 1st Place} & 1 / 23 & 8 / 23 & 3 / 23 & 15 / 23 \\
\bottomrule
\end{tabular}
}

% \vspace{-12pt}
\end{table}

\clearpage
\subsection{Experiments on various synthetic CO problems} \label{append:exp_general_co}

The experiments in this section cover various sample-efficient tasks in Euclidean and non-Euclidean combinatorial optimization.
Note that we assume the expensive black-box reward function in sample-efficient tasks.
In Euclidean CO tasks, the features of variables, such as their two-dimensional coordinates, satisfy Euclidean conditions (e.g., cost coefficients are defined as Euclidean distances). On the other hand, non-Euclidean CO problems lack these constraints, necessitating the encoding of higher-dimensional data, such as a distance matrix.

\subsubsection{Euclidean CO problems}
\paragraph{Experimental settings.}
We selected two representative routing tasks - the traveling salesman problem (TSP) and the capacitated vehicle routing problem (CVRP) with 50 and 100 customers.
The CVRP assumes multiple salesmen (i.e., vehicles) with limited carrying capacity; thus, if the capacity is exceeded, the vehicle must return to the depot. 
For base DRL methods, we employ the best-performing DRL methods, AM for TSP and Sym-NCO for CVRP. We follow the reported hyperparameters for the model in their original paper.\footnote{Sym-NCO: \url{https://github.com/alstn12088/Sym-NCO}}

\begin{table}
\centering
\caption{Experimental results on sample efficient Euclidean CO problems.}\label{tb:euclidean}

\begin{tabular}{llcccc}
\toprule
& & \multicolumn{2}{c}{$N=50$}& \multicolumn{2}{c}{$N=100$}\\
\cmidrule(lr{0.2em}){3-4}
\cmidrule(lr{0.2em}){5-6}
& Method& $K=200\text{K}$ &$K=2\text{M}$ &$K=200\text{K}$& $K=2\text{M}$\\
\midrule
% \multicolumn{8}{c}{Euclidean CO} \\
% \midrule
\parbox[t]{2mm}{\multirow{5}{*}{\rotatebox[origin=c]{90}{TSP}}} &
AM Critic & 6.541 $\pm$ 0.075
% & 6.242 $\pm$ 0.022
& 6.129 $\pm$ 0.021
& 9.600 $\pm$ 0.090
% & 9.097 $\pm$ 0.113
& 8.917 $\pm$ 0.115 \\ 
& AM Rollout 
& 6.708 $\pm$ 0.077 
% & 6.328 $\pm$ 0.029 
& 6.199 $\pm$ 0.014 
& 11.891 $\pm$ 1.008 
% & 9.736 $\pm$ 0.254 
& 9.193 $\pm$ 0.053 \\
& POMO 
& 7.910 $\pm$ 0.055 
% & 7.097 $\pm$ 0.007 
& 7.074 $\pm$ 0.010 
& 12.766 $\pm$ 0.358 
% & 11.019 $\pm$ 0.180 
& 10.964 $\pm$ 0.171 \\
& Sym-NCO
& 7.035 $\pm$ 0.209 
% & 6.436 $\pm$ 0.057 
& 6.334 $\pm$ 0.045 
& 10.776 $\pm$ 0.362 
% & 9.335 $\pm$ 0.089 
& 9.159 $\pm$ 0.056 \\
\cmidrule(lr{0.2em}){2-6}
&  \ours{} (ours) 
& \textbf{6.450 $\pm$ 0.053} 
% & \textbf{6.183 $\pm$ 0.034} 
& \textbf{6.038 $\pm$ 0.005} 
& \textbf{9.521 $\pm$ 0.098} 
% & \textbf{8.819 $\pm$ 0.028} 
& \textbf{8.573 $\pm$ 0.019} \\
\midrule
\parbox[t]{2mm}{\multirow{4}{*}{\rotatebox[origin=c]{90}{CVRP}}} &
AM Rollout & 
13.366 $\pm$ 0.199 
% & 12.247 $\pm$ 0.075 
& 11.921 $\pm$ 0.026 
& 23.414 $\pm$ 0.238 
% & 20.296 $\pm$ 0.802 
& 19.088 $\pm$ 0.232\\ 
& POMO & 
13.799 $\pm$ 0.310 
% & 12.731 $\pm$ 0.050 
& 12.661 $\pm$ 0.065 
& 22.939 $\pm$ 0.245 
% & 20.905 $\pm$ 0.382 
& 20.785 $\pm$ 0.403 \\
& Sym-NCO &
13.406 $\pm$ 0.204 
& 12.215 $\pm$ 0.124 
% & 11.964 $\pm$ 0.102 
& 21.860 $\pm$ 0.422 
% & 19.069 $\pm$ 0.138 
& 18.630 $\pm$ 0.106 \\
\cmidrule(lr{0.2em}){2-6}
&  \ours{} (ours) &  
\textbf{12.922 $\pm$ 0.071 }
% & \textbf{12.061 $\pm$ 0.116} 
& \textbf{11.721 $\pm$ 0.093} 
& \textbf{21.582 $\pm$ 0.149} 
% & \textbf{18.960 $\pm$ 0.067} 
& \textbf{18.304 $\pm$ 0.109} \\

\bottomrule
\end{tabular}

\end{table}

\begin{figure*}
\centering
         \includegraphics[width=0.47\textwidth]{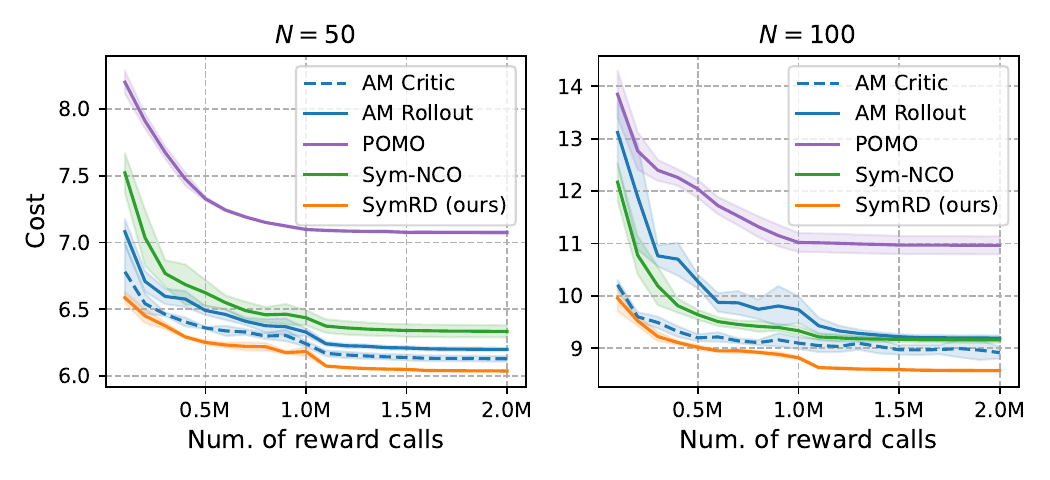}
        \includegraphics[width=0.47\textwidth]{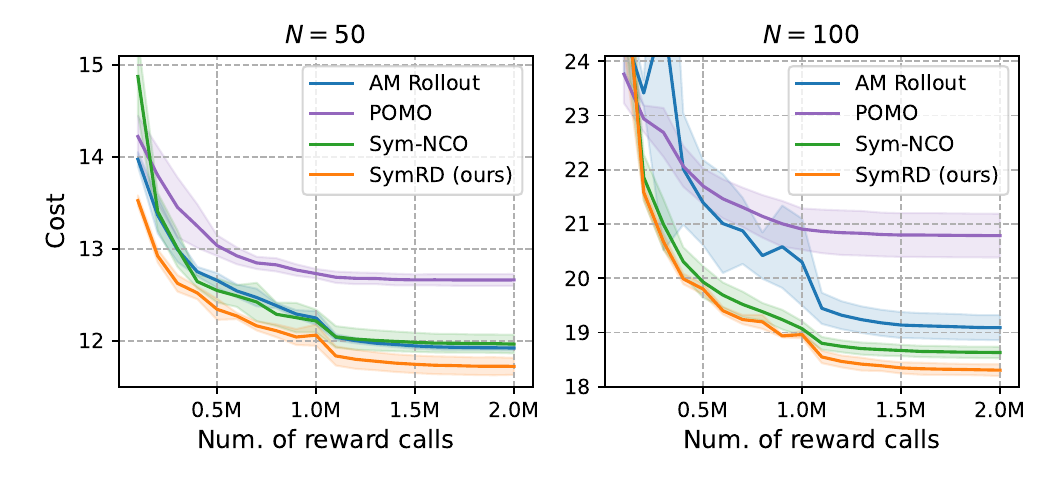}
\caption{Validation cost over computation budget on euclidean CO problems.}\label{figure:eu-co}
\end{figure*}

\paragraph{Results.}  The results in \cref{tb:euclidean} and \Cref{figure:eu-co} indicate that \ours{} consistently outperforms baseline methods in terms of achieving the lowest cost over the training budget. Note that ours employs the AM with critic baseline for TSP and Sym-NCO with the reduced number of augmentations for CVRP.
As depicted in \cref{tb:euclidean}, the most significant improvement over the base DRL models is observed in TSP100, with a percentage decrease of 3.86\%, and CVRP50, with a percentage decrease of 4.04\%.
While POMO and Sym-NCO consider the symmetric nature of CO, the required number of samples cancels out the benefits.
In contrast, our method utilizes the symmetric pseudo-labels generated via the training policy for free, enabling the policy to explore the symmetric space without increasing the number of required samples. As a result, \ours{} successfully improves sample efficiency.

\subsubsection{Non-Euclidean CO problems}
\paragraph{Experimental settings.}
Based on the work of \citet{kwon2021matrix}, we have selected two benchmark tasks, namely the asymmetric TSP (ATSP) and flexible flow-shop scheduling problems (FSSP). The ATSP is non-Euclidean TSP where the distance matrix could be non-symmetric, i.e., $\text{dist}(i,j) \neq \text{dist}(j,i)$, where $i$ and $j$ indicate cities. The FSSP is an important scheduling problem that assigns jobs to multiple machines to minimize total completion time.
As a baseline, we employ Matrix Encoding Network (MatNet) proposed to solve non-Euclidean CO.\footnote{MatNet: \url{https://github.com/yd-kwon/MatNet}} We compare ours with two versions of MatNet: MatNet-Fixed and MatNet-Sampled. MatNet-Fixed, the original version, explores $N$ heterogeneous starting points of trajectories, while MatNet-Sampled explores less than $N$ number of multiple trajectories with sampling strategy.

\begin{table}
\centering
\caption{Experimental results on sample efficient non-Euclidean CO problems.}\label{tb:non_euclidean}

\begin{tabular}{llcccc}
\toprule
& & \multicolumn{2}{c}{$N=50$}& \multicolumn{2}{c}{$N=100$}\\
\cmidrule(lr{0.2em}){3-4}
\cmidrule(lr{0.2em}){5-6}
& Method& $K=200\text{K}$ &$K=2\text{M}$ &$K=200\text{K}$ & $K=2\text{M}$ \\
\midrule
\parbox[t]{2mm}{\multirow{3}{*}{\rotatebox[origin=c]{90}{ATSP}}} &
MatNet-Fixed 
& 3.139 $\pm$ 0.024 
% & 2.128 $\pm$ 0.006 
& 2.000 $\pm$ 0.002 
& 4.400 $\pm$ 0.040 
% & 3.371 $\pm$ 0.024 
& 3.227 $\pm$ 0.016 \\
& MatNet-Sampled 
& 3.235 $\pm$ 0.021 
% & 2.250 $\pm$ 0.009 
& 2.019 $\pm$ 0.005 
& 4.324 $\pm$ 0.036 
% & 3.221 $\pm$ 0.044 
& 2.915 $\pm$ 0.040 \\ 
\cmidrule(lr{0.2em}){2-6}
&  \ours{} (ours)
& \textbf{2.845 $\pm$ 0.039} 
% & \textbf{2.125 $\pm$ 0.010} 
& \textbf{1.945 $\pm$ 0.003} 
& \textbf{3.771 $\pm$ 0.012} 
% & \textbf{2.712 $\pm$ 0.023} 
& \textbf{2.513 $\pm$ 0.022} \\
\midrule
\parbox[t]{2mm}{\multirow{3}{*}{\rotatebox[origin=c]{90}{FSSP}}} &
MatNet-Fixed 
& 56.350 $\pm$ 0.170
% & 55.588 $\pm$ 0.030 
& 55.341 $\pm$ 0.118
& 96.461 $\pm$ 0.206 
& 95.107 $\pm$ 0.072 \\
& MatNet-Sampled 
& 56.347 $\pm$ 0.234 
& 55.172 $\pm$ 0.032
& 96.256 $\pm$ 0.140 
& 94.978 $\pm$ 0.055 \\
\cmidrule(lr{0.2em}){2-6}
& \ours{} (ours)
& \textbf{56.104 $\pm$ 0.125}
% & 55.365 $\pm$ 0.126
& \textbf{55.110 $\pm$ 0.061}
& \textbf{96.030 $\pm$ 0.132}
& \textbf{94.934 $\pm$ 0.051} \\
\bottomrule
\end{tabular}

\end{table}

\begin{figure*}
\centering
         \includegraphics[width=0.47\textwidth]{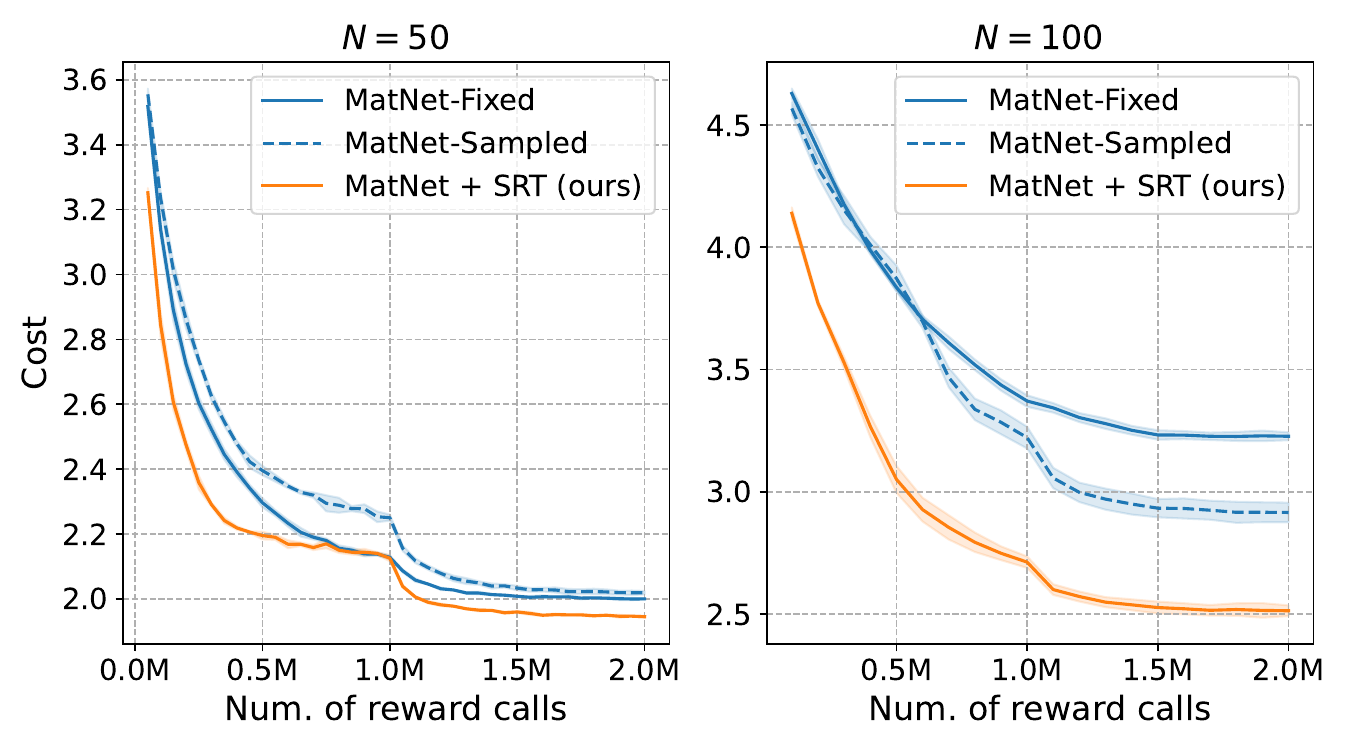}
        \includegraphics[width=0.47\textwidth]{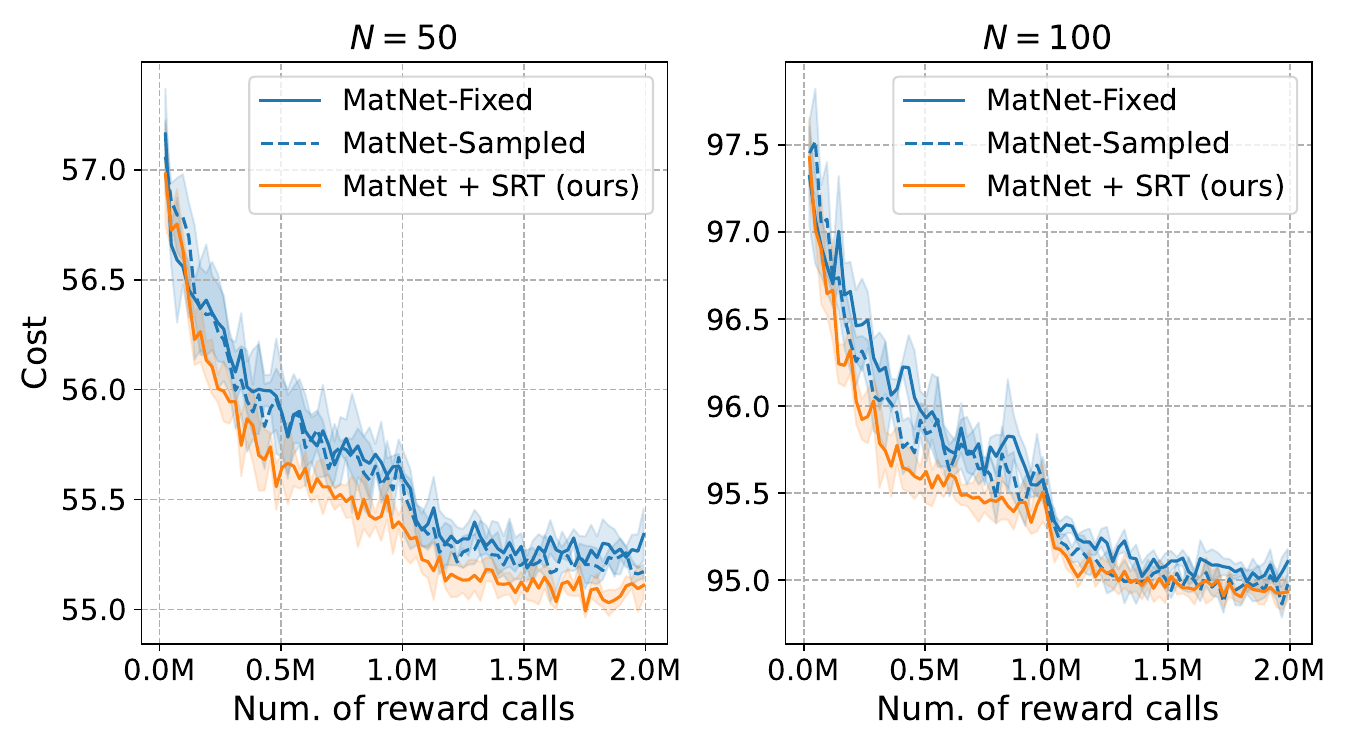}
\caption{Validation cost over computation budget on non-Euclidean CO problems.}
\label{figure:non_eu}
\end{figure*}

\textbf{Results.} The superior performance of \ours{} over MatNet-Fixed and MatNet-Sampled is demonstrated in both \cref{tb:non_euclidean} and \Cref{figure:non_eu}. We employ MatNet-Sampled as a base DRL method for both tasks and use the same number of multi-starting in ours and MatNet-Sampled. Notably, \ours{} outperforms MatNet-Sampled by a significant margin in the case of ATSP, with a performance gap of about 12\% at $N=100, K=200\text{K}$, where \ours{} achieves 3.771 and MatNet-Sampled achieves 4.324. 

\clearpage
\section{Further related works} \label{append:related}

\subsection{Deep reinforcement learning for combinatorial optimization} \label{append:drl}

Deep reinforcement learning (DRL) has emerged as a promising methodology for solving combinatorial optimization. Especially by selecting actions sequentially, i.e., in a constructive way, DRL policies generate a combinatorial solution. This approach is beneficial to producing feasible solutions that satisfy the complex constraints of CO by restricting action space using a masking scheme \citep{kool2018attention}. 
The foundational work of \citet{bello2017neural} introduced the actor-critic method for training PointerNet \citep{vinyals2015pointer} to solve TSP and the knapsack problem. Subsequently, several works were proposed to extend PointerNet into a Transformer-based model \citep{kool2018attention,xin2021multi} especially for routing problems. Building upon the success of the attention model \citep[AM; ][]{kool2018attention}, \citet{kwon2020pomo}, and \citet{kim2022sym} suggested enhanced reinforcement learning techniques by employing a precise baseline for REINFORCE based on symmetries in CO problems. 
On the other hand, various works have been suggested to solve broader ranges of CO problems \citep{khalil2017s2vdqn,ahn2020learning,zhang2020learning,jiang2021solving,kwon2021matrix,kim2021dpp,park2021schedulenet,zhang2023let,son2023equity}

Several studies have proposed to address challenges such as distributional shift and scalability \citep{hottung2021efficient,li2021l2d,ma2021hierarchical,choo2022simulation,qiu2022dimes,son2023meta,ma2023learning, jiang2023ensemble}.
It is noteworthy that besides the constructive approach, there is another stream, the improvement heuristic style \citep{hottung2020neural,xin2021neurolkh,kim2022nce,ye2023deepaco,kim2024ant}, though such studies fall outside our research scope.
Our research goal is to enhance the sample efficiency of constructive DRL methods for CO; the sample efficiency has been comparatively less explored in contrast to issues such as distributional shift and scalability in DRL for CO literature. This study offers an orthogonal but generally applicable approach to existing works in the field.

\subsection{Equivariant Deep Learning}

Equivariant DRL has also been extensively studied in recent years \cite{mondal2022eqr, van2020mdp, mondal2020group, wang2022so, deac2023equivariant}. This approach reduces search space by cutting out symmetric space using equivariant representation learning, such as employing equivariant neural networks \cite{cohen2016group, weiler2019general, satorras2021n}. 
Consequently, it leads to better generalization and sample efficiency.
Being different from these approaches, we focus on handling symmetries in decision space by exploring the symmetric space without restrictions on network structure. Therefore, employing equivariant DRL methods with our method is available when guaranteeing equivariance is crucial.

\subsection{Generative Flow Networks in Combinatorial Optimization}
Genetic Flow Networks or GFlowNets \citep{bengio2021gflownet} are suggested to sample discrete object $x$ from the target energy distribution, where $P(x) \propto e^{-\beta \mathcal{E}(x)}$. 
GFlowNets consider symmetric trajectories by modeling a sequential construction of combinatorial solutions as a flow on a directed acyclic graph, where each node corresponds to a state (a partial solution).
GFlowNets have been employed to solve various CO problems, whether when the reward evaluations are expensive or not.
\citet{bengio2021flow} firstly proposed GFlowNets with flow matching loss, a temporal difference-like loss. The work of \citet{bengio2021flow} is used to generate molecular graphs in de nove molecular optimization. Then, \citet{jain2022biological} introduced a model-based GFlowNet, called GFlowNet-AL, to generate biological sequences. 
In addition, \citet{zhang2023let} used GFlowNet to solve CO defined on graphs, like maximum independent set (MIS) problems.

On the other hand, several works have been proposed for better GFlowNet training, by introducing new loss functions \citep{bengio2021gflownet,malkin2022trajectory,madan2023sub-tb}, improved exploration strategies \citep{pan2023fl-gfn, jang2024learning}, experience replaying \citep{ma2023baking}, and credit assignment methods \citep{shen23guided_tb,pan2022generative,rector2023thompson,kim2023learning}.

\subsection{Black-box Combinatorial Optimization} \label{append:black-box}

% Our work is highly related to black-box optimization. Thus, we briefly introduce the related works for black-box combinatorial optimization.

Building on the great success of Bayesian optimization (BO) in black-box optimization \citep{bliek2023surrogate,irurozki2021unbalanced,lindauer2022smac3}, several works were suggested to apply BO to combinatorial decision.
Combinatorial Bayesian optimization solves a bi-level optimization where the upper problem is surrogate regression, and the lower problem is acquisition optimization. In the context of combinatorial space, the acquisition optimization is modeled as quadratic integer programming problems, which is NP-hard \citep{baptista2018bayesian,deshwal2022bayesian}. 

To address the NP-hardness of acquisition optimization, various techniques have been proposed, including continuous relaxation \citet{oh2019combinatorial}, sampling with simulated annealing \citep{deshwal2022bayesian}, genetic algorithms \citep{moss2020boss,tripp2021fresh}, and random walk explorer \citep{korovina2020chembo}. While these methods have demonstrated competitive performance in lower-dimensional tasks like neural architecture search (NAS), they often demand substantial computation time, particularly in molecular optimization tasks \citep{gao2022sample}.

As mentioned in \citep{deshwal2022bayesian} and \citep{gao2022sample}, one of the alternative approaches is to utilize a variational auto-encoder \citep[VAE; ][]{Kingma2014vae} to map the high-dimensional combinatorial space into ``compact'' continuous latent space to apply BO, like \citep{gomez2018automatic}. Though this approach has shown successful performance in molecular optimization, it also introduces variational error since VAE maximizes a lower bound of the likelihood, known as evidence lower bound (ELBO), not directly maximizes the likelihood.

%%%%%%%%%%%%%%%%%%%%%%%%%%%%%%%%%%%%%%%%%%%%%%%%%%%%%%%%%%%%%%%%%%%%%%%%%%%%%%%
%%%%%%%%%%%%%%%%%%%%%%%%%%%%%%%%%%%%%%%%%%%%%%%%%%%%%%%%%%%%%%%%%%%%%%%%%%%%%%%

\end{document}